\def\acceptedrevision{}
\def\manuscriptonly{}

\documentclass[sigconf, screen]{acmart}

\renewcommand\footnotetextcopyrightpermission[1]{}
\acmConference[]{\href{https://tetris-db.github.io}{tetris-db.github.io}\color{white}}{}{\color{black}}

\usepackage{xcolor}
\usepackage{colortbl}
\usepackage[normalem]{ulem}
\newif\ifrevisionmarkup
\ifdefined\acceptedrevision\revisionmarkupfalse\else\revisionmarkuptrue\fi
\DeclareRobustCommand{\revadd}[1]{#1}
\DeclareRobustCommand{\revpending}[2][]{#2}
\definecolor{revRoneColor}{RGB}{0,70,210}
\definecolor{revRsevenColor}{RGB}{0,135,60}
\definecolor{revRnineColor}{RGB}{120,40,180}
\definecolor{revAuthorColor}{RGB}{150,95,20}
\DeclareRobustCommand{\revRone}[1]{#1}
\DeclareRobustCommand{\revRseven}[1]{#1}
\DeclareRobustCommand{\revRnine}[1]{#1}
\DeclareRobustCommand{\revAuthor}[1]{#1}
\DeclareRobustCommand{\revdel}[1]{\ifrevisionmarkup\textcolor{red}{\sout{#1}}\fi}

\long\def\revdeletedtext#1{#1}
\newsavebox{\revremovedbox}
\newlength{\revremovedwidth}
\makeatletter
\long\def\revdelblock#1{\ifrevisionmarkup\par\if@noskipsec\leavevmode\par\fi\noindent\begingroup%
  \setlength{\revremovedwidth}{\linewidth}%
  \let\revdel\revdeletedtext\let\revadd\revdeletedtext\let\revpending\revdeletedtext\let\revdelblock\revdeletedtext%
  \let\revRone\revdeletedtext\let\revRseven\revdeletedtext\let\revRnine\revdeletedtext\let\revAuthor\revdeletedtext%
  \begin{lrbox}{\revremovedbox}\begin{minipage}{\revremovedwidth}\color{red}#1\end{minipage}\end{lrbox}%
  \begin{tikzpicture}%
  \node[inner sep=0pt,outer sep=0pt,anchor=south west] (removedtext) at (0,0) {\usebox{\revremovedbox}};%
  \draw[red] (removedtext.south west) -- (removedtext.north east);%
  \draw[red] (removedtext.north west) -- (removedtext.south east);%
  \end{tikzpicture}\par\endgroup\fi}
\makeatother
\usepackage{tikz}
\usepackage{xspace}
\usepackage[outline]{contour}
\contourlength{0.8pt}

\newif\ifshowautogen
\showautogenfalse %
\newcommand{\autogen}[1]{\ifshowautogen\textcolor{cyan}{#1}\else#1\fi}

\newcommand{\sys}[0]{Tetris\xspace}
\DeclareRobustCommand{\panellabel}[1]{%
    \tikz[baseline=(panelcaptionlabel.base)]{%
        \node[
            draw=black!75,
            fill=black!10,
            text=black,
            rounded corners=2pt,
            inner xsep=1pt,
            inner ysep=1pt,
            outer sep=0pt,
            text depth=0pt
        ] (panelcaptionlabel) {#1};%
    }%
}
\DeclareRobustCommand{\panellabelcolor}[2]{%
    \tikz[baseline=(panelcaptionlabel.base)]{%
        \node[
            draw=#1,
            fill=#1,
            text=white,
            rounded corners=4pt,
            inner xsep=1.7pt,
            inner ysep=1.5pt,
            outer sep=0pt,
            text depth=0pt
        ] (panelcaptionlabel) {#2};%
    }%
}

\usepackage{mathtools}
\usepackage{listings}
\usepackage{xcolor}

\definecolor{codegreen}{rgb}{0,0.6,0}
\definecolor{codegray}{rgb}{0.5,0.5,0.5}
\definecolor{codepurple}{rgb}{0.58,0,0.82}
\definecolor{backcolour}{rgb}{0.95,0.95,0.92}

\lstdefinestyle{codestyle}{
    backgroundcolor=\color{backcolour},   
    commentstyle=\color{codegreen},
    keywordstyle=\color{magenta},
    numberstyle=\tiny\color{codegray},
    stringstyle=\color{codepurple},
    basicstyle=\ttfamily\footnotesize,
    captionpos=b,
    keepspaces=true,
    showtabs=true,
    tabsize=2,
    frame=single,
    numbers=left
}
\renewcommand{\sectionautorefname}{\S\kern-1.5pt}
\renewcommand{\subsectionautorefname}{\S\kern-1.5pt}
\renewcommand{\subsubsectionautorefname}{\S\kern-1.5pt}

\usepackage{enumitem}
\setlist[itemize]{leftmargin=*, topsep=0.5em, itemsep=0pt, parsep=0pt}
\setlist[enumerate]{leftmargin=*, topsep=0.5em, itemsep=0pt, parsep=0pt}

\newcommand{\tileIrrelevanceMeanPct}{\autogen{94.6}}

\newcommand{\bboxOverheadPct}{\autogen{39}}

\newcommand{\comparePriorMeetFivePct}{\autogen{4}}
\newcommand{\comparePriorFailDatasetsFivePct}{\autogen{3}}

\newcommand{\compareAccuracyMatchedSpeedupMinFivePct}{\autogen{1.5}}
\newcommand{\compareAccuracyMatchedSpeedupMaxFivePct}{\autogen{17.4}}
\newcommand{\compareNaiveSpeedupMinFivePct}{\autogen{6.5}}
\newcommand{\compareNaiveSpeedupMaxFivePct}{\autogen{67.5}}

\newcommand{\compareAccuracyMatchedSpeedupMinTenPct}{\autogen{1.6}}
\newcommand{\compareAccuracyMatchedSpeedupMaxTenPct}{\autogen{24.7}}
\newcommand{\compareNaiveSpeedupMinTenPct}{\autogen{10.2}}
\newcommand{\compareNaiveSpeedupMaxTenPct}{\autogen{71.8}}
\newcommand{\compareMaxHotaImprovement}{\autogen{0.42}}
\newcommand{\compareMaxHotaImprovementFps}{\autogen{1000}}
\newcommand{\compareMaxHotaImprovementDataset}{\autogen{B3D4}}
\newcommand{\compareMatchedTputPriorConfigCount}{\autogen{54}}
\newcommand{\compareMatchedTputReachableCount}{\autogen{46}}
\newcommand{\compareMatchedTputHigherCount}{\autogen{45}}

\newcommand{\compareMatchedTputNotHigherMaxGap}{\autogen{0.001}}
\newcommand{\compareMatchedTputNotHigherDatasets}{\autogen{B3D1}}
\newcommand{\compareMatchedTputUnreachableCount}{\autogen{8}}
\newcommand{\compareMatchedTputUnreachableZeroHotaCount}{\autogen{7}}
\newcommand{\compareMatchedTputUnreachableRestDataset}{\autogen{B3D2}}
\newcommand{\compareMatchedTputUnreachableRestSpeedup}{\autogen{1.3}}
\newcommand{\compareMatchedTputUnreachableRestHotaDrop}{\autogen{0.20}}
\newcommand{\compareMatchedTputMedianHotaGain}{\autogen{0.09}}
\newcommand{\compareMatchedTputPerDatasetMaxGainMin}{\autogen{0.10}}

\newcommand{\ablationOneSpeedupMax}{\autogen{12.0}}

\newcommand{\ablationOneSpeedupAvg}{\autogen{6.9}}
\newcommand{\ablationTwoSpeedupMax}{\autogen{17.6}}

\newcommand{\ablationTwoSpeedupAvg}{\autogen{11.3}}
\newcommand{\ablationThreeSpeedupMax}{\autogen{67.5}}

\newcommand{\ablationThreeSpeedupAvg}{\autogen{26.7}}

\newcommand{\PtwozerofiveCompareClassifiersThreshold}{\autogen{.5}}

\newcommand{\PtwozerofiveCompareClassifiersBestDatasetDisplay}{\autogen{B3D 4}}

\newcommand{\PtwozerofiveCompareClassifiersBestFOneDeltaPp}{\autogen{8.60}}
\newcommand{\PtwozerofiveCompareClassifiersBestFOneModified}{\autogen{.84}}
\newcommand{\PtwozerofiveCompareClassifiersBestFOneBaseline}{\autogen{.76}}

\newcommand{\PtwozerofiveCompareClassifiersImprovedFOneDatasetCount}{\autogen{7}}
\newcommand{\PtwozerofiveCompareClassifiersDatasetCount}{\autogen{7}}

\newcommand{\PtwozerofiveCompareClassifiersMacroAvgFOneDeltaPp}{\autogen{7.15}}

\newcommand{\packingEfficacyBestValue}{\autogen{99.60}}
\newcommand{\packingEfficacyBestDatasetDisplay}{\autogen{B3D 3}}
\newcommand{\packingEfficacyWorstValue}{\autogen{63.02}}

\newcommand{\packingEfficacyMeanAcrossDatasets}{\autogen{91.86}}
\newcommand{\packingEfficacyDatasetCount}{\autogen{7}}

\newcommand{\packingEfficacyCalDoTOnePolyPerCanvas}{\autogen{15.2}}

\newcommand{\packingEfficacyCalDoTOnePolyMeanTiles}{\autogen{3.98}}

\newcommand{\packingEfficacyCalDoTTwoPolyPerCanvas}{\autogen{39.3}}

\newcommand{\packingEfficacyCalDoTTwoPolyMeanTiles}{\autogen{2.40}}

\newcommand{\packingEfficacyBThreeDTwoPolyPerCanvas}{\autogen{31.8}}

\newcommand{\packingEfficacyBThreeDTwoPolyMeanTiles}{\autogen{5.92}}

\newcommand{\packingEfficacyBThreeDThreePolyPerCanvas}{\autogen{61.3}}

\newcommand{\packingEfficacyBThreeDThreePolyMeanTiles}{\autogen{3.51}}

\newcommand{\mistrackStdCaldotoneYzerofiveBytetrackcythonTsSixzero}{\autogen{35.5}}

\newcommand{\mistrackStdJncsevenBytetrackcythonTsSixzero}{\autogen{18.2}}

\newcommand{\mistrackStdAllDatasets}{\autogen{20.6}}

\newcommand{\MistrackHeuristicMaxHotaLossPercent}{\autogen{8.27}}
\newcommand{\MistrackHeuristicAvgHotaLossPercent}{\autogen{0.78}}

\newcommand{\revisionDatasetCount}{\autogen{7}}
\newcommand{\revisionReferenceDetectorShareMinPct}{\autogen{99.4}}
\newcommand{\revisionReferenceDetectorShareMaxPct}{\autogen{99.9}}

\newcommand{\revisionClassifierFpsMin}{\autogen{436}}
\newcommand{\revisionClassifierFpsMax}{\autogen{1,261}}
\newcommand{\revisionDetectorFpsMin}{\autogen{14.5}}
\newcommand{\revisionDetectorFpsMax}{\autogen{32.9}}

\newcommand{\revisionPackingSparseNoPadRecallDropMin}{\autogen{3.2}}
\newcommand{\revisionPackingSparseNoPadRecallDropMax}{\autogen{9.9}}

\newcommand{\revisionPackingDeployedMaxChange}{\autogen{1.7}}

\newcommand{\revisionPackingSparsePadRecallMaxChange}{\autogen{0.5}}
\newcommand{\revisionPackingDenseMaxChange}{\autogen{1.4}}

\newcommand{\revisionLeapBThreeDVisibilityTenSeconds}{\autogen{3.8}}
\newcommand{\revisionLeapBThreeDVisibilityNinetySeconds}{\autogen{26.8}}
\newcommand{\revisionLeapCalDoTTwoMedianSamplingSeconds}{\autogen{4}}
\newcommand{\revisionLeapCalDoTTwoFewerThanTwoSamplesPct}{\autogen{59.7}}

\newcommand{\revisionBThreeDClipCount}{\autogen{20}}
\newcommand{\revisionBThreeDFrameCount}{\autogen{20,483}}
\newcommand{\revisionQueryAllLowerSecondGapPp}{\autogen{9.1}}
\newcommand{\revisionQueryLeapCases}{\autogen{21}}
\newcommand{\revisionQueryLeapBetterCount}{\autogen{15}}
\newcommand{\revisionNoScopeBThreeDStrictCostMin}{\autogen{11}}
\newcommand{\revisionNoScopeBThreeDStrictCostMax}{\autogen{22}}
\newcommand{\revisionQueryBestPct}{\autogen{81}}
\newcommand{\revisionQueryComparablePct}{\autogen{4.8}}
\newcommand{\revisionQueryWithinFivePct}{\autogen{14.3}}
\newcommand{\revisionQueryWithinFiveMaxGapPp}{\autogen{4.9}}

\newcommand{\revisionQueryAllBestPct}{\autogen{61.9}}

\newcommand{\revisionQueryAllWithinFivePct}{\autogen{28.6}}
\newcommand{\revisionQueryAllWithinFiveMaxGapPp}{\autogen{4.9}}

\newcommand{\revisionQueryAllLowerMaxGapPp}{\autogen{21.7}}
\newcommand{\revisionTetrisPreparationMinHours}{\autogen{6.1}}
\newcommand{\revisionTetrisPreparationMaxHours}{\autogen{31.9}}
\newcommand{\revisionOtifPreparationMinHours}{\autogen{2.5}}
\newcommand{\revisionOtifPreparationMaxHours}{\autogen{18.1}}
\newcommand{\revisionLeapPreparationMinHours}{\autogen{5.7}}
\newcommand{\revisionLeapPreparationMaxHours}{\autogen{12.0}}
\newcommand{\revisionTetrisValidationShareMinPct}{\autogen{77}}
\newcommand{\revisionTetrisValidationShareMaxPct}{\autogen{91}}

\newcommand{\revisionTetrisClassifierTrainingShareMinPct}{\autogen{4}}
\newcommand{\revisionTetrisClassifierTrainingShareMaxPct}{\autogen{12}}
\newcommand{\revisionTetrisTrainingDetectionShareMinPct}{\autogen{1}}
\newcommand{\revisionTetrisTrainingDetectionShareMaxPct}{\autogen{3}}

\newcommand{\revisionCrossTetrisReferenceMinHours}{\autogen{6.4}}
\newcommand{\revisionCrossTetrisReferenceMaxHours}{\autogen{32.4}}
\newcommand{\revisionSpeedupTetrisReferenceMin}{\autogen{6.5}}
\newcommand{\revisionSpeedupTetrisReferenceMax}{\autogen{67.5}}

\newcommand{\revisionCrossTetrisOtifMinHours}{\autogen{6.3}}
\newcommand{\revisionCrossTetrisOtifMaxHours}{\autogen{32.8}}
\newcommand{\revisionSpeedupTetrisOtifMin}{\autogen{1.5}}
\newcommand{\revisionSpeedupTetrisOtifMax}{\autogen{17.4}}

\newcommand{\revisionCrossLeapmatchedLeapMinHours}{\autogen{6.3}}
\newcommand{\revisionCrossLeapmatchedLeapMaxHours}{\autogen{35.9}}
\newcommand{\revisionSpeedupLeapmatchedLeapMin}{\autogen{4.3}}
\newcommand{\revisionSpeedupLeapmatchedLeapMax}{\autogen{40.4}}
\newcommand{\revisionCrossTetrisLatestHours}{\autogen{35.9}}

\newcommand{\revisionFixedClassifierPipelinePctMin}{\autogen{10.6}}
\newcommand{\revisionFixedClassifierPipelinePctMax}{\autogen{51.4}}
\newcommand{\revisionFixedClassifierDetectorPctMin}{\autogen{2.3}}
\newcommand{\revisionFixedClassifierDetectorPctMax}{\autogen{3.8}}

\newcommand{\revisionFixedClassifierSpeedupCapMin}{\autogen{26.4}}
\newcommand{\revisionFixedClassifierSpeedupCapMax}{\autogen{44.3}}
\newcommand{\revisionFixedSampledSpeedupCapMin}{\autogen{423}}
\newcommand{\revisionFixedSampledSpeedupCapMax}{\autogen{708}}
\newcommand{\revisionFixedPruningPipelinePctMin}{\autogen{1.6}}
\newcommand{\revisionFixedPruningPipelinePctMax}{\autogen{4.4}}
\newcommand{\revisionFixedPruningNaivePctMin}{\autogen{0.28}}
\newcommand{\revisionFixedPruningNaivePctMax}{\autogen{0.66}}
\newcommand{\revisionFixedPackingPipelinePctMin}{\autogen{1.8}}
\newcommand{\revisionFixedPackingPipelinePctMax}{\autogen{3.3}}
\newcommand{\revisionFixedPackingNaivePctMin}{\autogen{0.14}}
\newcommand{\revisionFixedPackingNaivePctMax}{\autogen{1.08}}
\newcommand{\revisionFixedRenderingPackingPctMin}{\autogen{64.5}}
\newcommand{\revisionFixedRenderingPackingPctMax}{\autogen{83.9}}

\newcommand{\revisionFixedIlpVariablesMin}{\autogen{1,603}}
\newcommand{\revisionFixedIlpVariablesMax}{\autogen{9,165}}
\newcommand{\revisionFixedIlpConstraintsMin}{\autogen{6,815}}
\newcommand{\revisionFixedIlpConstraintsMax}{\autogen{51,063}}
\newcommand{\revisionFixedClassifierDetectorSpeedupMin}{\autogen{26}}
\newcommand{\revisionFixedClassifierDetectorSpeedupMax}{\autogen{44}}
\newcommand{\revisionRgbClassifierDetectorAvgPct}{\autogen{1.9}}
\newcommand{\revisionAugmentedClassifierDetectorAvgPct}{\autogen{3.3}}
\newcommand{\revisionFixedIlpRuns}{\autogen{200}}

\newcommand{\revisionFixedIlpBudgetSeconds}{\autogen{0.1}}

\newcommand{\revisionFixedIlpInputTilesShort}{\autogen{1.4M}}
\newcommand{\revisionFixedIlpPrunedTilesShort}{\autogen{562k}}

\newcommand{\revisionFixedIlpWithinSixtySeconds}{\autogen{152}}
\newcommand{\revisionFixedIlpWithinOneHour}{\autogen{166}}
\newcommand{\revisionFixedIlpHitOneHour}{\autogen{34}}

\newcommand{\revisionFixedIlpHourAllTilesReducedShort}{\autogen{2.2k}}
\newcommand{\revisionFixedIlpHourAllTilesReducedPct}{\autogen{0.4}}
\newcommand{\revisionFixedIlpHourBoundPct}{\autogen{0.5}}
\newcommand{\revisionFixedIlpWithinPointOneSecond}{\autogen{97}}
\newcommand{\revisionFixedIlpWithinOneSecond}{\autogen{120}}
\newcommand{\revisionFixedIlpWithinTenSeconds}{\autogen{137}}

\newcommand{\revisionResolutionRatioMin}{\autogen{15.6}}
\newcommand{\revisionResolutionRatioMax}{\autogen{21.0}}

\newcommand{\revisionPaddingSparseNoPadCallsMin}{\autogen{4.2}}
\newcommand{\revisionPaddingSparseNoPadCallsMax}{\autogen{7.5}}
\newcommand{\revisionPaddingSparsePadCallsMin}{\autogen{12.3}}
\newcommand{\revisionPaddingSparsePadCallsMax}{\autogen{25.0}}
\newcommand{\revisionPaddingDenseNoPadCallsMin}{\autogen{12.4}}
\newcommand{\revisionPaddingDenseNoPadCallsMax}{\autogen{29.8}}
\newcommand{\revisionPaddingDensePadCallsMin}{\autogen{45.2}}
\newcommand{\revisionPaddingDensePadCallsMax}{\autogen{91.2}}

\begin{document}

\title{\sys{}: Tile-level Sampling for Efficient and High-Fidelity Video Object Tracking}

\newcommand{\berkeley}{%
  \affiliation{%
    \institution{U. of California, Berkeley}%
    \city{Berkeley}%
    \state{California}%
    \country{USA}%
  }%
}

\settopmatter{authorsperrow=4}
\makeatletter
\author@bx@sep=1pt\relax %
\makeatother

\author{Chanwut Kittivorawong}
\email{chanwutk@berkeley.edu}
\orcid{0000-0002-2884-2221}
\berkeley

\author{Alena Chao}
\email{alenachao@berkeley.edu}
\orcid{0009-0005-5005-6197}
\berkeley

\author{Charlie Si}
\email{charliesi@berkeley.edu}
\orcid{0009-0006-9594-6262}
\berkeley

\author{Alvin Cheung}
\email{akcheung@cs.berkeley.edu}
\orcid{0000-0001-6261-6263}
\berkeley

\begin{abstract}
Track materialization converts raw videos into reusable object tracks that downstream queries can run against without rerunning tracking,
but extracting those tracks efficiently and with high fidelity remains expensive.
Prior systems reduce track materialization cost through temporal frame sampling,
but aggressive sampling spaces each track's detection points too far apart
to faithfully capture the object's actual trajectory.
In stationary video, however,
large portions of each frame contain no objects of interest,
and different sampling rates can be used to extract tracks from the remaining regions.
Leveraging this idea, we present \sys{},
a \revAuthor{high-throughput} track-extraction system that decomposes videos into a tile-based \emph{polyomino} data model,
enabling fine-grained spatiotemporal pruning that reduces detector calls with minimal fidelity loss.
\sys{} implements track materialization in three steps: 
first, a classifier identifies relevant tiles and groups them into polyominoes.
Then, \revAuthor{it uses} an integer linear program (ILP) to prune redundant polyominoes under a user-specified accuracy constraint,
before packing the remaining polyominoes into canvases to minimize detector calls.
Across \revisionDatasetCount{} stationary-video datasets,
\sys{} stays within a 5\% tracking accuracy loss as compared to a reference pipeline that processes every frame in its entirety,
while prior systems exceed this bound on \comparePriorFailDatasetsFivePct{}
of the \revisionDatasetCount{} datasets.
Moreover, with this 5\% bound,
\sys{} achieves
up to \compareAccuracyMatchedSpeedupMaxFivePct{}$\times$ higher throughput
than prior systems,
and up to \compareNaiveSpeedupMaxFivePct{}$\times$ higher than the reference pipeline.
Conversely,
\sys{} delivers up to \compareMaxHotaImprovement{} higher HOTA tracking accuracy
than the best prior system at a matched throughput \revAuthor{of \compareMaxHotaImprovementFps{} FPS}.
\end{abstract}

\maketitle

\section{Introduction}
\label{sec:introduction}

Track materialization converts raw
videos into a reusable set of object tracks
that downstream analytics can query repeatedly~\cite{bastani2022otif,kossmann2023vetl}.
For stationary cameras deployed in traffic monitoring and surveillance,
each track is a per-object trajectory that can be used in queries such as counting,
retrieval,
and motion analysis.
Because each materialized track can serve many downstream queries,
systems aim to compute tracks
only
once,
then
store
and reuse them.

The bottleneck in this workload is track extraction,
the compute-intensive process that produces reusable tracks from raw video.
Modern tracking-by-detection
pipelines spend \revisionReferenceDetectorShareMinPct{}--\revisionReferenceDetectorShareMaxPct{}\% of their runtime inside the object detector%
,
and long continuous video streams amplify this cost at scale.
Reducing the number of detector invocations,
or the amount of pixel data each invocation must process,
is therefore the primary lever for improving throughput.

Many downstream tasks require fine-grained trajectories,
not merely object presence.
For instance, decentralized vehicle coordination in understructured traffic scenes
relies on per-vehicle trajectories recovered from object tracks~\cite{wu2025b3d},
and traffic-flow studies such as the I-24 MOTION phantom-jam experiment depend on recovering when each vehicle decelerates,
re-accelerates,
and whether its motion propagates or damps a stop-and-go wave~\cite{manke2022phantomjams}.
Missing, imprecise, or temporally coarse tracks are harmful during abrupt motion changes;
producing tracks at the fidelity that such queries require is exactly what makes track extraction expensive.

Prior track-materialization systems reduce cost through two strategies,
both incurring accuracy losses.
OTIF~\cite{bastani2022otif} and LEAP~\cite{xu2024leap} use whole-frame sampling;
with aggressive frame sampling,
the resulting tracks contain detection points spaced too far apart
to faithfully capture fine-grained motion such as abrupt decelerations.
OTIF additionally crops the sampled frames with axis-aligned rectangular regions of interest (ROIs),
that is,
rectangles whose edges are parallel to the image axes.
Such windows can still enclose irrelevant pixels when the underlying relevant region is non-rectangular.
\autoref{sec:track-materialization} discusses both strategies in detail.

Stationary-camera deployments are common in traffic monitoring and surveillance,
and they expose a property that prior work has not fully exploited:
{\em since the camera does not move and the scene's physical layout is fixed,
the spatial distribution of relevant content is stable over time.}
Objects recur in the same image regions
and follow the same scene-induced motion patterns.
This stability makes it possible to prune at a granularity finer than whole frames
and with shapes finer than rectangular ROIs.

We identify three properties of stationary video that make this stability exploitable,
to be discussed in~\autoref{sec:motivation}:

\noindent
\emph{\bf Spatial Irrelevance}: across our \revisionDatasetCount{} evaluation datasets,
a mean of \tileIrrelevanceMeanPct\% of spatial regions per frame contain no objects of interest.

\noindent
\emph{\bf Spatially Varying Sampling}: different sampling rates can be used for different regions,
depending on their motion patterns,
making any single whole-frame sampling rate either wasteful or inaccurate.

\noindent
\emph{\bf Non-rectangular \revAuthor{Relevant Regions}}: relevant regions form connected
but non-rectangular shapes whose axis-aligned enclosing rectangles contain on
average \bboxOverheadPct\% more area than the \revAuthor{relevant} regions themselves,
measured across our \revAuthor{seven} evaluation datasets.

To exploit these properties, we present \sys, a track-extraction system that decomposes stationary-camera video into a tile-based \emph{polyomino} data model~\cite{golomb1996polyominoes}, enabling fine-grained spatiotemporal pruning that reduces detector calls with minimal fidelity loss.
\sys{} divides each frame into a grid of square tiles and groups connected relevant tiles into polyominoes,
non-rectangular regions that tightly enclose relevant content (\autoref{sec:data-model}).
Importantly,
\sys{} is agnostic to the object detector and tracker,
and it lets the user define their accuracy target (\autoref{sec:pareto-frontier}).
Based on the target, \sys's execution engine (\autoref{sec:exec}) runs an optimized \emph{tracking pipeline}
of three operators (classify, prune, pack) upstream of the user-provided object detector,
then unpacks the resulting detections for the tracker (\autoref{sec:overview}).
This extraction-time pipeline is preceded by a one-time training phase that,
once per dataset,
trains a lightweight relevance classifier specialized to the user-provided detector (\autoref{sec:specialization}) and learns the maximum tile sampling gaps the pruner relies on (\autoref{sec:learn-constraints}).
\revRnine{Here,
a dataset is a large collection of videos from a single stationary camera.
The learned artifacts are specific to the dataset and the user-provided detector and tracker;
changing the dataset,
detector,
or tracker requires
rerunning the reference pipeline and repeating the necessary classifier training,
sampling-gap profiling,
and Pareto frontier generation.}

To use \sys{},
users provide a stationary-camera video dataset,
an object detector and a tracker,
and either a throughput requirement or an accuracy requirement.
\sys{} then runs a one-time preparation step on samples from the dataset,
producing the \emph{learned artifacts} needed for efficient extraction
(a relevance classifier and maximum tile sampling gaps)
and a Pareto frontier of candidate configurations that helps the user select an operating point.
After the user selects a configuration,
\sys's execution engine extracts the tracks on the remaining video data using that configuration,
and the extracted tracks are returned to the user.

On \revisionDatasetCount{}
stationary-video datasets~\cite{kang2017noscope,bastani2022otif,wu2025b3d},
\sys{} keeps the loss in HOTA~\cite{luiten2021hota} tracking accuracy
below 5\% relative to the reference pipeline (\autoref{sec:tracking-by-detection}),
while prior systems exceed this bound on \comparePriorFailDatasetsFivePct{} of the \revisionDatasetCount{} datasets.
Furthermore, with this 5\% bound,
\sys{} delivers \compareAccuracyMatchedSpeedupMinFivePct{}--\compareAccuracyMatchedSpeedupMaxFivePct{}$\times$ higher throughput
than prior systems,
and \compareNaiveSpeedupMinFivePct{}--\compareNaiveSpeedupMaxFivePct{}$\times$ higher throughput than the reference pipeline.
With a 10\%
loss
bound,
\sys{} achieves \compareAccuracyMatchedSpeedupMinTenPct{}--\compareAccuracyMatchedSpeedupMaxTenPct{}$\times$ higher throughput than prior systems
and \compareNaiveSpeedupMinTenPct{}--\compareNaiveSpeedupMaxTenPct{}$\times$ higher throughput than the reference pipeline
(\autoref{sec:evaluation}).

In sum, we make the following contributions.
\begin{itemize}
    \item A \emph{tile-based polyomino data model and an execution engine} for
    efficient
    track extraction.
    The engine runs three operators upstream of the user-provided detector:
    a classifier identifies polyominoes of relevant tiles,
    an ILP prunes redundant polyominoes constrained by learned maximum tile sampling gaps,
    and a polyomino packing algorithm packs the remaining tiles into a minimum number of canvases
    to reduce object detector invocations.
    \item A \emph{method for learning maximum tile sampling gaps}
    from empirical per-tile mistrack rates.
    For each mistrack-rate tolerance,
    \sys{} selects the largest measured gap that remains within the tolerance at each tile,
    turning an exponential search over per-tile gap assignments into a one-dimensional sweep over the scalar tolerance.
    At runtime,
    these gaps become ILP constraints that let the execution engine prune polyominoes according to empirically measured mistrack rates.
    \item An evaluation on seven stationary-video datasets using an accuracy (HOTA) or throughput constraint. 
    With a 5\% HOTA-loss bound,
    \sys{} meets the bound on all \revAuthor{datasets}, while prior systems fail to do so on \comparePriorFailDatasetsFivePct{} of the \revisionDatasetCount{} datasets. Compared with prior systems with the 5\% bound,
    \sys{} delivers up to \compareAccuracyMatchedSpeedupMaxFivePct{}$\times$ higher throughput.
    And, with the same throughput bound as prior systems, 
    \sys{} achieves up to \compareMaxHotaImprovement{} higher HOTA
    at \compareMaxHotaImprovementFps{} FPS.
\end{itemize}

\section{Background and Observations}
\label{sec:preliminaries}

We first discuss the tracking-by-detection pipeline that \sys accelerates
(\autoref{sec:tracking-by-detection}),
position our work within efficient track materialization
(\autoref{sec:track-materialization})
\revRseven{and other video analytics systems
(\autoref{sec:related-systems}),}
and present observations on stationary video that motivate \sys{}'s design
(\autoref{sec:motivation}).

\subsection{Tracking by Detection}
\label{sec:tracking-by-detection}

Tracking by detection is the most widely used paradigm for multi-object tracking~\cite{luo2021mot}.
As shown in \autoref{fig:track-by-detect},
such pipeline takes an $N$-frame video as input and processes it in two stages.
First,
an object detector~\cite{ren2016fasterrcnn,redmon2016yolo,lin2017focal} runs independently on each of the $N$ frames,
producing a set of bounding boxes per frame,
\revAuthor{where each box $(x_1, y_1, x_2, y_2)$ gives the pixel coordinates of a rectangle enclosing a detected object.}
Let $N'$ denote the number of detection calls.
In this naive pipeline, $N' = N$ since the detector is invoked once per frame.
Second,
an object tracker performs \emph{data association};
it links the bounding boxes across consecutive frames that likely correspond to the same physical object,
assigning each a consistent track identifier to produce a set of tracks~\cite{bewley2016sort,zhang2022bytetrack,cao2023ocsort,wojke2018reid,wojke2017deepsort}.
Each resulting track is a sequence of per-frame bounding boxes
sharing a track identifier.

While the object detector utilizes expensive machine learning inference on every frame,
data association can be performed using simple statistical methods such as bounding box overlap~\cite{bewley2016sort,zhang2022bytetrack,cao2023ocsort}.
Hence, detection dominates the pipeline's computational cost and accounts for \revAuthor{\revisionReferenceDetectorShareMinPct{}--\revisionReferenceDetectorShareMaxPct{}\% of the detection and tracking compute} of an unoptimized
full-frame, every-frame tracking-by-detection pipeline,
measured across the 7 evaluation datasets of \autoref{sec:evaluation}.
Reducing the number of detector invocations
is therefore the primary strategy for improving end-to-end throughput.
We call this unoptimized full-frame, every-frame pipeline built from the user-provided detector and tracker the \emph{reference pipeline}; \sys{} optimizes against it and we use it throughout as the accuracy and runtime baseline.

\subsection{Track Materialization}
\label{sec:track-materialization}

Track materialization is the data management task of converting raw video
into persistent object tracks that downstream queries can reuse, 
and \sys accelerates the track-extraction step that produces these reusable tracks from stationary camera streams.

The primary cost of track materialization is producing those reusable tracks.
As
detection dominates runtime,
prior systems reduce this cost by reducing detector computation.
One common approach is temporal frame sampling.
OTIF~\cite{bastani2022otif} samples frames uniformly to reduce execution time proportionally,
and LEAP~\cite{xu2024leap} introduces a predictive strategy that skips frames whose object states can be inferred from previously sampled frames.
A complementary approach is spatial pruning within each sampled frame.
For example, OTIF uses a proxy model to identify axis-aligned regions of interest (ROIs) that the detector then processes.
Both approaches reduce detector computation.
Temporal sampling reduces the number of frames passed to the detector,
and spatial pruning reduces the amount of pixel content the detector must process within sampled frames.
But each has a corresponding limitation.
Frame sampling erases inter-frame motion that fine-grained tracking requires,
and rectangular ROIs waste detector work when relevant contents are not perfectly rectangular.
We evaluate track extraction quality with HOTA~\cite{luiten2021hota},
which jointly penalizes missing detections,
fragmented tracks,
and identity switches,
making it the appropriate fidelity metric for the fine-grained tracking we target.
\autoref{sec:motivation} quantifies the performance gap these limitations cause on stationary videos.

\subsection{%
\revRseven{%
Specialization in Video Analytics Systems}%
}
\label{sec:related-systems}

\revRseven{Video analytics systems share optimization techniques but specialize them to different tasks,
scene assumptions,
and pipeline stages.
Studying them across tasks helps identify their strengths.
Systems with compatible assumptions can complement one another by optimizing different pipeline components.
This variety lets users
choose the most specialized system that still supports their
queries.}

\revRseven{%
RegenHance~\cite{wang2025regenhance} groups selected macroblocks into connected regions resembling polyominoes,
then packs their bounding boxes across frames for super-resolution.
MOSAIC~\cite{gokarn2023mosaic} uses motion and tracking to select rectangular regions of varying sizes,
then resizes and packs regions from concurrent streams into detector inputs.}

\revRseven{\sys{} shares these grouping and packing building blocks,
but targets track materialization.
\sys combines the tile and polyomino data model with an execution engine that operates on these units for joint spatial and temporal selection of detector input.
Guided by observations from stationary video (\autoref{sec:motivation}),
we remove irrelevant content,
group the remainder tightly,
and retain regions at different frequencies.
Keeping less content reduces detector work but risks missing detections or breaking associations.
To manage this tradeoff,
we learn each location's sensitivity to sampling and validate configurations to balance throughput against tracking fidelity.}

\revRseven{In multi-device settings,
systems optimize offloading and camera coordination.}
\revRseven{Glimpse~\cite{chen2015glimpse} selects whole cached frames for local tracking and triggers server recognition;
PreActo~\cite{nguyen2023preacto} predicts cross-camera handovers.}
\revRseven{In video compression,}
\revRseven{How2Compress~\cite{wu2025how2compress} reduces encoded bytes through macroblock quantization,
while Crucio~\cite{zhu2024crucio} combines frame filtering,
learned compression,
and decoding optimizations.
\sys{} complements these \revRseven{coordination and} codec optimizations by pruning and packing decoded content to reduce the remaining detector computation.}

\revRseven{OTIF and LEAP remain our closest object-level materialization baselines (\autoref{sec:track-materialization}).
NoScope~\cite{kang2017noscope} uses frame-difference filters to reject predicted empty frames or reuse presence labels.
Seiden~\cite{bang2023seiden} allocates whole-frame samples with a bandit strategy guided by variation in sampled query results,
then interpolates labels for unsampled frames.
These frame-level labels do not by themselves provide per-object tracks.
\sys{} prunes within frames,
which allows sampling different regions at different frequencies
while still materializing detailed tracks for reuse across downstream queries.}

\subsection{Observations from Stationary Video\revAuthor{s}}
\label{sec:motivation}

We present empirical evidence from our seven stationary-video datasets
for the three properties claimed in \autoref{sec:introduction}.
We formally describe these datasets in \autoref{sec:evaluation};
here,
we use two of them as running examples:
CalDoT1,
a highway traffic camera from the OTIF benchmark~\cite{bastani2022otif},
and B3D4,
an intersection camera dataset~\cite{wu2025b3d}.

\textbf{\em Observation 1: Spatial Irrelevance.}
\autoref{fig:observations}~\panellabel{Left} shows the percentage of frames in which each tile location contains a detected object
in CalDoT1.
Most tile locations are rarely relevant,
and many are never relevant.
Across our
\revAuthor{7}
datasets from different traffic cameras,
a mean of \tileIrrelevanceMeanPct\% of tiles per frame contain no objects of interest,
quantifying the opportunity for tile-level filtering.
\revRnine{A small image classifier over fixed-size tiles
can identify them \revisionFixedClassifierDetectorSpeedupMin{}--\revisionFixedClassifierDetectorSpeedupMax{}$\times$ faster than the detector running on the same frames (\autoref{sec:ilp-cost}).}

\textbf{\em Observation 2: Spatially Varying Sampling.}
\autoref{fig:observations} \panellabel{Middle} visualizes the per-tile mistrack percentage
on the B3D4 dataset when sampling every four frames.
In the B3D4 intersection scene,
tiles along the approaching lanes (right lanes) of each leg exhibit higher mistrack rates than tiles along the exiting lanes (left lanes).
This is because vehicles approaching the intersection accelerate and decelerate unpredictably,
making their positions harder for the tracker to interpolate across skipped frames.
Tiles immediately before the stop line are an exception because vehicles are nearly stationary and thus easy to track.
Vehicles exiting the intersection maintain steady speed,
allowing the tracker to predict their positions accurately even with aggressive frame skipping.
These spatial divergences in tracking difficulty are reflected in the high variance of per-tile mistrack rates.
Across the seven datasets combined, the standard deviation is \mistrackStdAllDatasets\%.
On individual datasets, the rates vary significantly as well;
for example, the standard deviation is \mistrackStdJncsevenBytetrackcythonTsSixzero\% on B3D4
and \mistrackStdCaldotoneYzerofiveBytetrackcythonTsSixzero\% in CalDoT1.
These spatial patterns show that using a single whole-frame sampling rate is suboptimal,
and that a system that learns maximum tile sampling gaps
can significantly reduce the number of tiles processed with minimal tracking accuracy loss.

\textbf{\em Observation 3: \revAuthor{Non-rectangular Relevant Regions}.}
\autoref{fig:observations} \panellabel{Right} compares the area of a connected relevant-tile group
against the area of its axis-aligned window in B3D4.
On average across our datasets,
axis-aligned windows contain \bboxOverheadPct\% more tiles
than the connected groups they enclose.
This overhead compounds with the pruning from Observations~1 and~2, since a system that prunes at the granularity of rectangular regions
reintroduces irrelevant content that tile-level pruning would have discarded.

\textbf{\em From observations to system design.}
These three observations motivate the core components of \sys:
tile-level relevance classification (\autoref{sec:relevance-classification}, Obs.~1),
per-tile polyomino pruning, which skips more tiles in regions that tolerate infrequent observation (\autoref{sec:prune}, Obs.~2),
and polyomino grouping rather than rectangular ROIs (\autoref{sec:data-model}, Obs.~3).

\begin{figure}[t]
  \centering
  \includegraphics[width=\columnwidth]{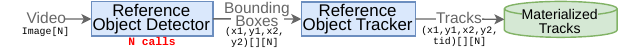}

  \caption{Tracking-by-detection pipeline.}
  \Description{A pipeline diagram showing video frames flowing into an object detector, producing bounding boxes, then into an object tracker producing tracks, and finally into materialized storage.}
  \label{fig:track-by-detect}

\end{figure}

\begin{figure}[t]
    \centering
    \includegraphics[width=.297\columnwidth]{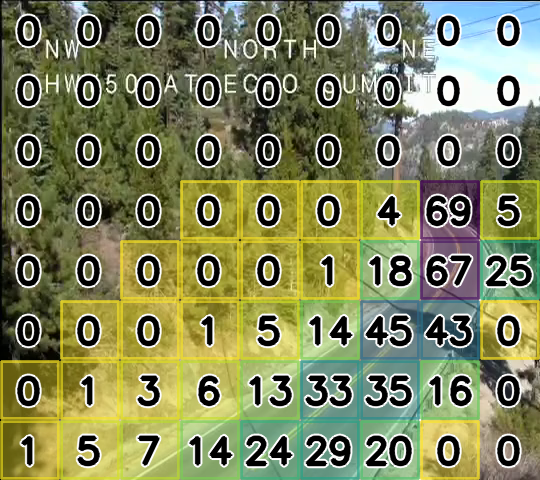}\hfill
    \includegraphics[width=.396\columnwidth]{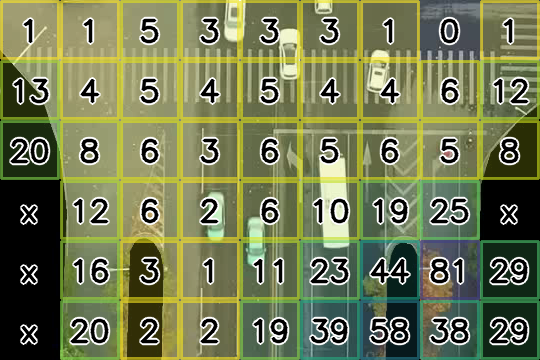}\hfill
    \includegraphics[width=.297\columnwidth]{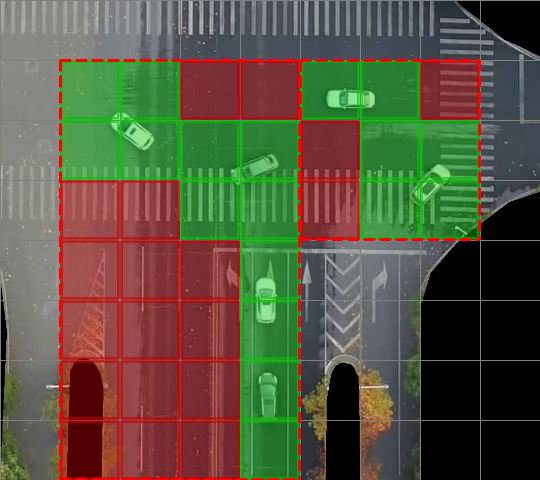}

    \Description{}
    \caption{\panellabel{Left} Per-tile relevance percentage across CalDoT1.
    Most tile locations are rarely or never relevant.
    \panellabel{Middle} Per-tile mistrack rate (\%) for B3D4 when sampling every 4 frames.
    \panellabel{Right}
    A connected group of relevant tiles (green tiles)
    and its axis-aligned window (red dotted lines).
    Red tiles are irrelevant but included in the window.}
    \label{fig:observations}

\end{figure}

\section{\sys{} Overview}
\label{sec:overview}

\begin{figure*}
  \includegraphics[width=\textwidth]{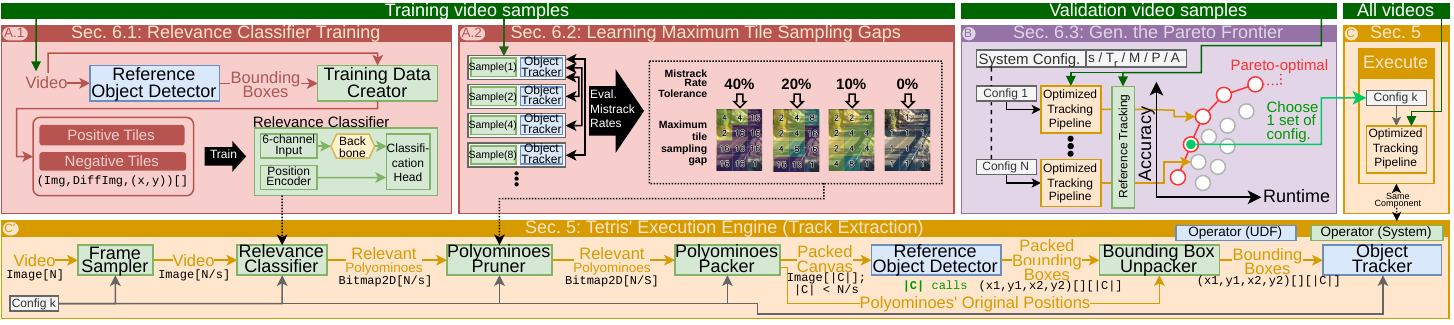}

  \caption{System overview.
    The learning stage produces the learned artifacts:
    the relevance classifier (\autoref{sec:specialization})
    and maximum tile sampling gaps (\autoref{sec:learn-constraints}).
    Pareto frontier generation produces a throughput-accuracy curve from swept system configurations (\autoref{sec:pareto-frontier}).
    The user selects a configuration,
    then the \sys{} execution engine (\autoref{sec:exec})
    uses it to extract tracks on the full dataset.
    UDF operators are supplied by the user (the detector, and optionally the tracker);
    System operators are \sys{}'s optimization operators.}
  \Description{}
  \label{fig:system}

\end{figure*}

Consider a user with a large video dataset captured by stationary cameras
and a user-provided object detector and tracker for identifying and tracking objects of interest
(e.g.,
vehicles in traffic monitoring).
The user's goal is to materialize the tracks for all objects of interest across the dataset,
subject to a throughput 
or accuracy requirement.
\sys accelerates track extraction under either requirement,
maximizing throughput subject to a bound on accuracy loss
or maximizing accuracy subject to a throughput floor%
.

To achieve this goal, \sys first produces
the \emph{learned artifacts}
(a relevance classifier and maximum tile sampling gaps)
and a \emph{throughput--accuracy frontier} that helps the user choose an operating point;
only then does it extract the \emph{final tracks}
by applying the selected configuration to the remaining video data.
The artifacts and the frontier guide extraction,
but they are not the final tracks.
\sys does not change how tracks are represented or queried downstream;
it changes how efficiently they are extracted.

At its core,
\sys divides each video frame into a grid of square \emph{tiles}
and treats tiles as the fundamental unit of processing.
Each tile is classified as either relevant (containing objects of interest)
or irrelevant (background),
allowing \sys to discard large portions of a frame
without running the expensive object detector.
Connected relevant tiles within a frame form a \emph{polyomino},
an arbitrary two-dimensional shape composed of edge-adjacent tiles,
which must be processed together to avoid splitting objects across tile boundaries.
Not every polyomino needs to be processed in every frame.
Our learned \emph{maximum tile sampling gaps}
determine how frequently each region must be sampled
to keep tracking accuracy within a given tolerance.
After pruning,
the remaining polyominoes are packed into rectangular \emph{canvases} for batch object detection.
We formally define tiles and polyominoes in \autoref{sec:data-model}.

Shown in \autoref{fig:system},
\sys learns the artifacts and generates the Pareto frontier on small samples,
then extracts tracks on the full dataset with a user-selected configuration.
\sys samples two disjoint video subsets from the input video dataset:
a training sample for learning the relevance classifier and maximum tile sampling gaps
and a validation sample for generating the Pareto frontier.
The sample sizes are user-settable;
our evaluation uses at most 60 one-minute videos per sample (\autoref{sec:evaluation}).
Once the user chooses a configuration from the generated Pareto frontier,
\sys{} then extracts tracks efficiently from the remaining data via the \sys{} execution engine (\autoref{sec:exec}).
\revRnine{This workflow is agnostic to the choice of object detector and tracker.
Its learned artifacts and throughput--accuracy frontier are specific to each dataset and detector/tracker combination,
and \sys{} regenerates them when that combination changes.
\revRnine{On our
datasets,
this one-time preparation pays off within \revisionCrossTetrisLatestHours{} hours and amortizes over further videos from the same distribution (\autoref{sec:preparation-cost}).}}

\textbf{\em Learned Artifact Construction (\autoref{sec:specialization}, \autoref{sec:learn-constraints}).}
Using the training sample,
\sys trains a lightweight relevance classifier specialized for the user-provided detector (\autoref{sec:specialization})
and learns maximum tile sampling gaps (\autoref{sec:learn-constraints}).
The user-provided detector and tracker together define the unoptimized full-frame,
every-frame \emph{reference pipeline} (\autoref{fig:track-by-detect}).
Because the input video lacks ground-truth tracks,
\sys uses this pipeline's tracking results as the target for accuracy evaluation.
The relevance classifier learns to predict whether a tile contains objects of interest
from the tile's color image,
the difference between consecutive frames,
and the tile's position.
The maximum tile sampling gaps specify,
for each tile location,
how frequently that location must be sampled to maintain accurate tracks.

\textbf{\em Pareto Frontier Generation (\autoref{sec:pareto-frontier}).}
Using the validation sample,
\sys runs the unoptimized reference pipeline to establish the reference tracks,
then \sys's execution engine extracts tracks from the validation sample
under a range of
configurations (\autoref{tab:frontier-sweep}).
Each configuration produces a
throughput-accuracy operating point.
The output is a set of Pareto-optimal configurations
that lie on the throughput-accuracy frontier,
each representing a
tradeoff
between throughput and tracking accuracy against the reference tracks.
\begin{table}[t]
  \centering
  \footnotesize
  \caption{System configuration parameters.}

  \label{tab:frontier-sweep}
  \begin{tabular}{@{}ll@{}}
    {\bf Parameter} & {\bf Candidate values considered} \\
    \midrule
    Whole-frame sampling rate~$s$
      & $1$, $2$, $4$, $8$, $16$
      (every $s$-th frame; \autoref{sec:additional-knobs}) \\
    Relevance threshold~$T_r$ (\autoref{sec:relevance-classification})
      & $0.25$, $0.5$, $0.75$ \\
    Mistrack-rate tolerance~$\bar{M}$ (\autoref{sec:prune})
      & none,
      \revAuthor{$0.2$, $0.4$, $0.6$}
      \\
    Tile padding~$P$ (\autoref{sec:pack})
      & $0$;
      $0.5$ tile (top-\revAuthor{right});
      $0.5$ tile (bottom-\revAuthor{left});
      $1$ tile \\
    Tracker option~$A$ (\autoref{sec:track})
      & user-provided tracker, SORT \\
  \end{tabular}

\end{table}

\textbf{\em Track Extraction (\autoref{sec:exec}).}
Given the user's throughput or accuracy requirement,
the user selects a configuration from the Pareto frontier,
and \sys's execution engine extracts tracks on the remaining video data
by executing
a tracking pipeline
of three \sys{} operators upstream of the user-provided detector,
followed by a lightweight unpacking step before the tracker.
First, the relevance classifier evaluates each tile
and groups connected relevant tiles into polyominoes (\autoref{sec:relevance-classification}).
Second,
an ILP prunes polyominoes based on the maximum tile sampling gaps to
minimize the total number of tiles processed (\autoref{sec:prune}).
Third,
a two-dimensional first-fit-descending algorithm packs the surviving polyominoes
into rectangular canvases of the same resolution as the original video frames (\autoref{sec:pack}).
The user-provided object detector then processes each canvas,
and \sys{} unpacks the resulting bounding boxes back to original frame coordinates
for the tracker (\autoref{sec:detect-unpack-track}).

\paragraph{\revRone{Specialization and scope.}}
\revRone{%
\sys{}
tracks objects from
videos on a single server without distributed camera coordination.
Its learned artifacts are specific to each dataset and detector/tracker pair.
Changes to the dataset,
video distribution,
or either model require repeating the affected learning and validation%
.
We train,
validate,
and test within each stationary-camera dataset (\autoref{sec:evaluation});
cross-camera transfer and robustness to weather or traffic shifts are not evaluated.}

\section{Data Model for Optimization}
\label{sec:data-model}

\sys operates on a tile-level data model
that lets its execution engine process only regions relevant to object tracking.
This section defines its 2 core abstractions:
tiles and polyominoes.
\revRseven{Both are units of optimization for faster tracking
rather than
the system's output.}

\subsection{Tiles and Relevance}
\label{sec:dm-tiles}

Each video frame is divided into a grid of square tiles of size $TS \times TS$ pixels.
For a frame of height $H_F$ and width $W_F$, the grid contains $H = \lfloor H_F/TS \rfloor$ rows and $W = \lfloor W_F/TS \rfloor$ columns;
\revAuthor{frames
with
sizes
not divisible by $TS$ are resized}
\revAuthor{to fit the grid}.
We denote the tile at row $i$ and column $j$ for a
particular
frame by its position $(i, j)$.

\sys{} classifies each tile as either \emph{relevant} or \emph{irrelevant} for object tracking.
Relevant tiles visually contain the objects of interest (e.g., vehicles in traffic monitoring),
while irrelevant tiles contain only background content such as empty roads,
trees,
or sky, as shown in \autoref{fig:tile-relevance} (A).
Formally, a relevance classifier assigns each tile a score
$Score_{i,j} \in [0, 1]$,
where values close to~1 indicate high relevance and values close to~0 indicate irrelevance.
A tile is relevant if $Score_{i,j} \geq T_r$ for a threshold~$T_r$;
otherwise it is irrelevant.
The threshold~$T_r$ is a system configuration parameter,
set by the configuration the user selects after Pareto frontier generation (\autoref{sec:pareto-frontier}).
Irrelevant tiles are discarded entirely, saving \sys{} from running the expensive object detector on regions that contain no objects of interest.
The design and training of the relevance classifier are described in \autoref{sec:relevance-classification}.

\revRone{\sys{}
uses
$TS$=60.}
\revRone{Smaller tiles prune irrelevant content more finely but increase the number of tiles,
and thus the classification
time,
while larger tiles retain more irrelevant background pixels.}

\subsection{Relevant Polyominoes}
\label{sec:dm-polyominoes}

\begin{figure}[t]
    \centering
    \includegraphics[width=.9\columnwidth]{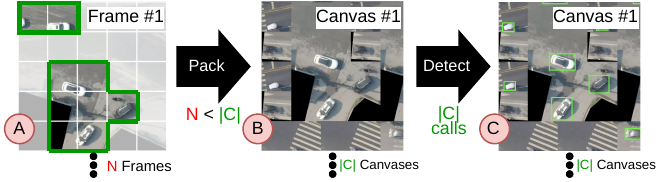}

    \Description{}
    \caption{A video frame divided into a grid of tiles.
    Relevant tiles (highlighted) contain objects of interest,
    while irrelevant tiles (grayed) are discarded to save computation.
    Connected relevant tiles form a polyomino (green border).
    Polyominoes from multiple frames are packed into a single canvas.
    }
    \label{fig:tile-relevance}    

    \vspace{-1pt}
\end{figure}

An object of interest may span multiple adjacent tiles within the same frame.
If each relevant tile were processed independently by the object detector,
the detector could incorrectly localize the object, for instance,
reporting two half-detections for a vehicle that straddles two tiles.
Therefore, connected relevant tiles must be processed together, as shown in \autoref{fig:tile-relevance} (A).

We group connected relevant tiles into a \emph{polyomino}~\cite{golomb1996polyominoes,knuth2000dancinglinks},
an arbitrary shape formed by edge-connected tiles on a grid.
Let $P$ denote the list of \revAuthor{disjoint} relevant polyominoes in the frame.
Each polyomino $p_k \in P$ ($1 \le k \le |P|$) is a set of tile positions on the grid
and $|p_k|$ denotes the number of tiles in $p_k$.
\revRseven{A polyomino is a unit for joint pruning and packing and may contain multiple objects.
It does not represent an object segmentation mask or identity.}

Unlike prior systems such as OTIF~\cite{bastani2022otif} that group relevant tiles into the smallest enclosing rectangular region of interest (ROI),
polyominoes can be of arbitrary shape.
This flexibility is important because rectangular ROIs can be highly wasteful.
For example, if objects align diagonally across a frame,
the smallest enclosing rectangle may span the entire frame,
forcing the system to process large irrelevant regions.
Polyominoes avoid this excess coverage by tightly following the shape of the relevant area.
The procedure for constructing polyominoes from classified tiles is detailed in \autoref{sec:output-polyominoes}.

In subsequent sections, when discussing multiple frames, we add a frame index $f$ to these symbols (e.g., $Score_{f,i,j}$, $P_f$, $p_{f,k}$).

\section{\sys{} Execution Engine}
\label{sec:exec}

Once the user selects a system configuration
\revRnine{(\autoref{tab:frontier-sweep})},
\sys's execution engine extracts tracks on the remaining video data.
The engine consumes the two learned artifacts constructed in \autoref{sec:e2e},
the relevance classifier and the maximum tile sampling gaps;
this section treats them, along with the selected configuration, as given.
The goal is to reduce time needed to materialize tracks while maintaining tracking accuracy.
At a high level,
the execution engine identifies the regions of each frame that are relevant to tracking,
and packs them into canvases that reduce the number of calls to the object detector,
as shown in \autoref{fig:system} \panellabelcolor{orange}{C$'$}.

The execution engine runs an optimized tracking pipeline
composed of \revAuthor{3} \sys{} operators upstream of the user-provided detector,
followed by a lightweight unpacking step before the tracker.
The detector must be user-provided because it encodes domain-specific knowledge of what objects look like;
the tracker only consumes per-frame bounding boxes,
so the user may provide a tracker or use \sys{}'s built-in default, SORT~\cite{bewley2016sort}.
Before the \revAuthor{3} operators run,
the engine applies standard
whole-frame sampling,
processing only every $s$-th frame ($s = 1$ retains every frame).
The rate~$s$ is set by the user-selected configuration \revRnine{(\autoref{tab:frontier-sweep})}.
The \revAuthor{3} \sys{} operators are:
\begin{enumerate}
\item
{\bf Relevance Classification (\autoref{sec:relevance-classification})} identifies the regions in video frames that contain objects of interest.
A video frame is divided into a grid of square tiles.
Tiles that contain objects of interest are marked as relevant.
The rest are marked as irrelevant.
Connected relevant tiles within the same frame form a polyomino.
The operator outputs a list of relevant polyominoes.

\item
{\bf Polyomino Pruning (\autoref{sec:prune})} samples at the tile level rather than frame level.
Given the maximum tile sampling gaps learned offline,
it prunes polyominoes whose tiles are
covered by neighboring frames,
minimizing the
number of tiles to process.

\item
{\bf Polyomino Packing (\autoref{sec:pack})} packs the remaining polyominoes into canvases to minimize the number of
detector calls.
\end{enumerate}
After these three operators,
\sys{} invokes the user-provided detector on each packed canvas,
unpacks the resulting detections back to their original frame coordinates,
and feeds them to the selected tracking algorithm to produce object tracks
(\autoref{sec:detect-unpack-track}).

\subsection{Relevance Classification}
\label{sec:relevance-classification}

Prior work~\cite{bastani2022otif} in video analytics systems uses coarse-grained proxy models to identify regions that
contain objects of interest,
avoiding calls to the detector on regions predicted to be irrelevant.
Unlike these prior proxy models,
which classify regions from pixel values alone,
\sys{}'s relevance classifier additionally conditions on per-tile motion (frame differences) and spatial position,
which is
informative for stationary videos.
Frame differences highlight
motion that is usually absent from background tiles,
while tile position captures stable scene priors such as lanes, sidewalks, and sky.
Together,
these cues help determine tile relevance in visually ambiguous regions where RGB appearance alone leaves useful evidence unused.

\paragraph{Model Architecture}
Building on the observations above,
we note two additional cues for tile-level classification:
the frame-difference image is a strong indicator of relevance (moving objects produce visible local changes),
and tile position provides a stable spatial prior (e.g., sky tiles are rarely relevant).
Our classifier therefore takes the tile's RGB image,
the frame-difference image,
and the tile position as input,
and
\revRnine{determines if the tile contains objects of interest to the user-provided detector.}
We use ShuffleNet~\cite{ma2018shufflenet}
as the backbone,
and extend it to
additionally take in
the difference between consecutive frames
and tile positions.

We now formalize the classifier's inputs,
its per-tile scoring function,
and its per-frame output.
At video frame $f$, the tile position is represented as a 2D coordinate $(i, j)$.
Let $\mathrm{Img}_{f,i,j}$ denote the RGB tile image at position $(i, j)$ in frame $f$,
and $\Delta\mathrm{Img}_{f,i,j}$ the corresponding frame-difference image, defined as
    \revAuthor{$\Delta\mathrm{Img}_{f,i,j} = \lvert \mathrm{Img}_{f,i,j} - \mathrm{Img}_{f-1,i,j} \rvert$,}
where $\lvert \cdot \rvert$ denotes element-wise absolute value.

Our classification function on each tile is defined as
\begin{equation}
    \label{eq:classify}
    Score_{f,i,j} = \mathrm{Relevance}(\mathrm{Img}_{f,i,j}, \Delta\mathrm{Img}_{f,i,j}, (i, j))
\end{equation}
The score is a scalar value between 0 and 1,
with 1 indicating high relevance and 0 indicating irrelevance.
If $Score_{f,i,j}$ is greater than a threshold $T_r$,
the tile at position $(i, j)$ in frame $f$ is considered relevant.
Otherwise,
the tile is considered irrelevant.

For a particular frame $f$,
\sys{} outputs a matrix $Scores_f$ of size $H \times W$, where
each element $Score_{f,i,j}$ is the relevance score for the tile at position $(i, j)$ in frame $f$.

\begin{figure}[t]
    \centering
    \includegraphics[width=\columnwidth]{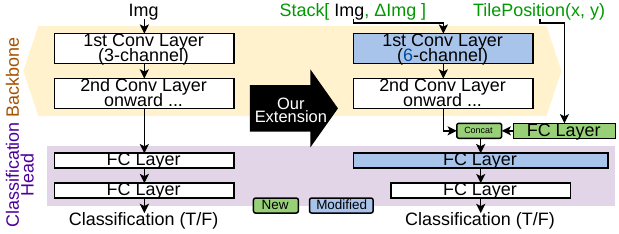}

    \Description{}
    \caption{\revAuthor{\panellabel{Left} An off-the-shelf image classifier.
    \panellabel{Right} Our relevance classifier architecture.}}
    \label{fig:classifier}

\end{figure}

The architecture of %
our
classifier is shown in \autoref{fig:classifier}.
The first layer of the backbone of an off-the-shelf classifier is a convolutional layer that accepts the color image (3 channels).
Here, we replace the first layer with a 6-channel convolutional layer that accepts the color image and the difference between consecutive frames.
Accepting the image frame difference allows the model to capture the motion of the objects in the video.
In the classification head,
we concatenate the feature output from the backbone (visual and motion features) with the tile position embeddings,
and pass the concatenated output through two fully connected layers to obtain the relevance score for each tile.
The position embeddings are computed through a learned fully connected layer
instead of by directly passing the tile position to the classification head.
The final output is a scalar value between 0 and 1, where 1 means the tile is highly relevant, and 0 means the tile is irrelevant.

\paragraph{Output Polyominoes}
\label{sec:output-polyominoes}
After classifying all tiles in a frame,
\sys{} groups connected relevant tiles into polyominoes as defined in \autoref{sec:dm-polyominoes}.
The resulting polyominoes cannot be passed to the object detector directly due to their irregular shapes;
\autoref{sec:pack} describes how \sys{} packs them into rectangular canvases for batch detection.

\subsection{Polyomino Pruning}
\label{sec:prune}

Prior work in frame sampling~\cite{kang2017noscope,xu2024leap,bastani2022otif,li2020reducto,bastani2020miris,bang2023seiden}
either includes all relevant regions in a video frame or
excludes the whole video frame from processing.
This is wasteful because some regions of the video frame may require more frequent sampling to track objects accurately.
For example,
as observed in \autoref{sec:motivation},
tiles along approaching lanes at an intersection exhibit higher mistrack rates
because vehicles accelerate and decelerate unpredictably,
making their positions harder for the tracker to interpolate across skipped frames.
Conversely,
vehicles exiting the intersection maintain steady speed,
allowing the tracker to predict their positions accurately even with aggressive frame skipping.
Therefore, we develop a novel sampling method that performs tile-level
rather than frame-level sampling.

\subsubsection{Maximum Tile Sampling Gaps}
To sample at the tile level,
\sys{} bounds how long each tile location may go unsampled%
.
Given a mistrack-rate tolerance $0 \leq \bar{M} \leq 1$ \revRnine{(\autoref{tab:frontier-sweep})},
\sys{} derives a matrix $\bar{G}^{\bar{M}}$ of maximum tile sampling gaps, of size $H \times W$ over the tile grid,
as illustrated in \autoref{fig:system} \panellabelcolor{red}{A.2}.
Here,
$(i, j)$ indexes a tile location in a frame's $H \times W$ tile grid (\autoref{sec:dm-tiles}).
Each element $\bar{g}^{\bar{M}}_{i,j}$ is a positive integer
that states the maximum gap between consecutive samples
of tile at location $(i, j)$ before the tracker will exceed the
set mistrack tolerance rate of $\bar{M}$.
Our optimization objective is hence to minimize the number of tiles to sample,
while adhering to the maximum tile sampling gaps $\bar{G}^{\bar{M}}$.
\autoref{sec:learn-constraints} describes how \sys{} obtains $\bar{G}^{\bar{M}}$
\revRnine{based on the per-tile mistrack rates measured empirically%
.}

\subsubsection{Optimization Problem}
\begin{figure}[t]
    \centering
    \resizebox{\columnwidth}{!}{\begin{tikzpicture}[x=0.5cm,y=0.54cm]
  \tikzstyle{labelstyle}=[font=\footnotesize]
  \tikzstyle{guidestyle}=[line width=0.2pt, draw=black!18]
  \tikzstyle{dropstyle}=[draw=black!30, fill=black!8, rounded corners=0.5pt, line width=0.3pt]
  \tikzstyle{keepstyle}=[draw=teal!70!black, fill=teal!25, rounded corners=0.5pt, line width=0.45pt]
  \tikzstyle{countstyle}=[font=\footnotesize, text=teal!70!black]

  \foreach \x/\name/\c in {0/A/4,1/B/4,2/C/2,3/D/2,4/E/2,5/F/2,6/G/4,7/H/4} {
    \node[labelstyle, font=\scriptsize] at (\x+0.5,1.05) {$\name$};
    \node[labelstyle, font=\scriptsize] at (\x+0.5,0.55) {$\c$};
    \draw[guidestyle] (\x,0.2) -- (\x,-1.77);
    \draw[guidestyle] (\x+1,0.2) -- (\x+1,-1.77);
  }
  \node[labelstyle, font=\scriptsize, anchor=east] at (-0.25,1.05) {position};
  \node[labelstyle, font=\scriptsize, anchor=east] at (-0.25,0.55) {$\bar{g}_j^{\bar{M}}$};

  \node[labelstyle, rotate=90, anchor=south] at (-1.2,-0.8) {Frame};
  \foreach \y/\t in {-0.05/f_1,-0.35/f_2,-0.65/f_3,-0.95/f_4,-1.25/f_5,-1.55/f_6} {
    \node[labelstyle, font=\tiny, anchor=east] at (-0.25,\y) {\scalebox{0.8}{$\t$}};
  }

  \draw[keepstyle] (0,-0.17) rectangle (3,0.07);
  \draw[dropstyle] (5,-0.17) rectangle (7,0.07);

  \draw[keepstyle] (2,-0.47) rectangle (6,-0.23);

  \draw[dropstyle] (1,-0.77) rectangle (3,-0.53);
  \draw[keepstyle] (5,-0.77) rectangle (8,-0.53);

  \draw[keepstyle] (2,-1.07) rectangle (6,-0.83);

  \draw[keepstyle] (0,-1.37) rectangle (3,-1.13);
  \draw[dropstyle] (6,-1.37) rectangle (8,-1.13);

  \draw[keepstyle] (3,-1.67) rectangle (7,-1.43);

  \node[labelstyle, font=\tiny, anchor=west] at (8.25,0.4) {Total = 27 tiles};
  \node[countstyle, font=\tiny, anchor=west] at (8.25,0.0) {Selected total = 21 tiles};
  \node[labelstyle, font=\tiny, anchor=west, align=left] at (8.25,-0.6) {$(F$--$G)$ in $f_1$ dropped:\\
  $F$, $G$ covered by $f_2$, $f_3$};
  \node[labelstyle, font=\tiny, anchor=west, align=left] at (8.25,-1.4) {similarly, $(B$--$C)$ in $f_3$\\
  and $(G$--$H)$ in $f_5$ dropped};
\end{tikzpicture}}

    \Description{}
    \caption{A 1D illustration of polyomino pruning with heterogeneous maximum tile sampling gaps.
    Each row shows one frame's polyominoes;
    teal boxes are kept and gray boxes are pruned.
    As a result of pruning,
    21 out of 27 tiles are kept.}
    \label{fig:pruning-example}

\end{figure}
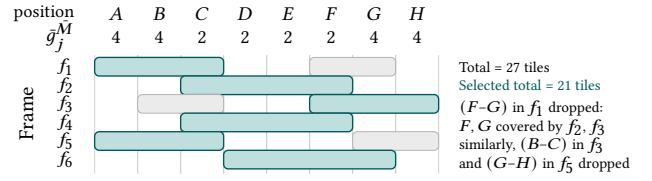

Sampling tiles is not trivial because we cannot simply consider each tile to be independent of the others.
As explained in \autoref{sec:output-polyominoes}, connected relevant tiles must be processed together; they cannot be split apart.
If a tile is sampled, the whole polyomino that contains the tile must be sampled.
This problem then becomes choosing a subset of relevant polyominoes to process,
such that the total number of tiles to sample is minimized,
while strictly adhering to the maximum tile sampling gaps $\bar{G}^{\bar{M}}$.
\autoref{fig:pruning-example} shows a simple 1D example.
Although the actual problem is two-dimensional,
the 1D abstraction makes the coupling between per-position maximum tile sampling gaps and polyomino-level selection easier to see.
In the example,
each frame contains one or two polyominoes of varying shapes,
and per-tile maximum tile sampling gaps range from 2 to 4.
The solver exploits the fact that
some polyominoes' positions are already covered by neighboring frames:
for instance,
positions $F$ and $G$ in frame $f_1$ are covered
by the polyominoes kept in $f_2$ and $f_3$,
so the solver drops the polyomino $(F$--$G)$ while retaining the polyomino $(A$--$B$--$C)$.
This partial-frame pruning reduces the total
from 27 tiles (frame-level selection) to 21,
a saving that frame-level approaches cannot achieve.

\subsubsection{ILP Formulation}
In general, solving this problem is NP-hard (proof in \autoref{sec:np-pruning}).
Therefore, we formulate it as an integer linear program
and rely on a modern ILP solver,
which handles \sys{}'s instance sizes efficiently in practice. Specifically,
assume that we select the subset of relevant polyominoes from $N$ video frames.
\begin{itemize}
\item
Let $P$ be the set of all relevant polyominoes in all $N$ frames.
\item
Let $P_f$ be the list of relevant polyominoes in frame $f$.
\item
Let $p_{f,k}$ be the $k$-th relevant polyomino in frame $f$.
\item
Let $s_{f,k}$ be a binary variable that is 1 if the polyomino $p_{f,k}$ is selected, and 0 otherwise.
\item
Let $Size(p_{f,k})$ be the number of tiles in the polyomino $p_{f,k}$.
\item
Let $Covers(p_{f,k},i,j)$ be a Boolean function that returns $true$
if the polyomino $p_{f,k}$ covers the tile at position $(i, j)$.
Formally,%
\[\revAuthor{Covers(p_{f,k},i,j) \iff (i,j) \in p_{f,k}}.\]%
\item
Let $CoverageSpan(f_{st},f_{en},i,j)$ be the set of binary variables $s_{f,k}$
for corresponding polyominoes that cover tile location $(i, j)$
in frames $f_{st}$ through $f_{en}$.
Formally,
\begin{equation*}
CoverageSpan(f_{st}, f_{en}, i, j) \coloneqq
\left\{\;
s_{f,k}
\;\middle|\;
\begin{aligned}
&f_{st} \le f \le f_{en},
1 \le k \le |P_f|,\\
&Covers(p_{f,k},i,j)
\end{aligned}
\;\right\}.
\end{equation*}
As a shorthand, we define the coverage span with size $\bar{g}^{\bar{M}}_{i,j}$ starting at frame $f$
and tile at position $(i, j)$ as
\begin{equation*}
CS(\bar{M},f,i,j) \coloneqq
CoverageSpan(f,f+\bar{g}^{\bar{M}}_{i,j}-1,i,j).
\end{equation*}
\end{itemize}

We can then represent the problem as a set of constraints as follows.
For each tile at position $(i,j)$,
each coverage span of size $\bar{g}^{\bar{M}}_{i,j}$ must either contain at least one selected polyomino or be empty:
\begin{equation*}
   \label{eq:constraint-ilp}
   \begin{aligned}
        \forall i\, \forall j\, \forall f \in [1,\, N - \bar{g}^{\bar{M}}_{i,j} + 1]:\quad
        &
        ( 1 \le \textstyle \sum\limits_{\mathclap{s \in CS(\bar{M},f,i,j)}} s )
        \;\lor\;
        (CS(\bar{M},f,i,j)=\emptyset)
    \end{aligned}
\end{equation*}
This enforces that each tile position $(i,j)$ has a sampling gap no greater than $\bar{g}^{\bar{M}}_{i,j}$ frames.
The quantification ranges only over windows that fit within the $N$-frame horizon.
A window extending past frame $N$ imposes no constraint,
since fewer than $\bar{g}^{\bar{M}}_{i,j}$ frames remain.
However,
empty spans are permitted for tiles that are not relevant for $\bar{g}^{\bar{M}}_{i,j}$ consecutive frames,
\revAuthor{preventing infeasible ILPs.}

Given the constraints,
the objective of the ILP problem is to minimize the number of selected tiles,
which is:
\begin{equation*}
    \label{eq:objective-ilp}
    \textstyle \sum\limits_{f=1}^{N} \sum\limits_{k=1}^{|P_f|} s_{f,k} \times Size(p_{f,k})
\end{equation*}

The ILP outputs a set of selected polyominoes $P^s$,
defined as
\begin{equation*}
\label{eq:result-ilp}
P^s \coloneqq
\left\{
p_{f,k}
\;
|
\;
s_{f,k} = 1
\right\}.
\end{equation*}
\revRnine{%
The objective minimizes the number of selected tiles
while satisfying the maximum tile sampling gaps.
\sys{}
limits solving to 0.1\,s and uses
its best feasible solution,
which need not be optimal.}

\revRnine{To keep the ILP size independent of video length,
\sys{} solves a separate ILP for each
1-minute
window of video.
A longer video adds windows,
so pruning cost grows linearly with video length.%
}

\subsection{Polyomino Packing}
\label{sec:pack}

After pruning,
the surviving polyominoes must be converted into detector-compatible inputs.
Our key idea is to pack multiple polyominoes,
potentially from different frames,
into fixed-size 2D canvases.
Each canvas has the same resolution as the original video frame,
so one detector invocation can process several relevant regions at once.
This reduces the number of detector calls without changing the detector itself.
Keeping the original frame resolution also preserves object scale,
avoiding the scale mismatch that rescaling would introduce for the user-provided detector.

\subsubsection{Optimization Problem}
Given a set of relevant polyominoes $P = \{p_1, p_2, \ldots, p_n\}$,
we aim to pack them into 2D canvases of size $H \times W$ tiles,
equivalent to the original frame size of $H_F \times W_F$ pixels.
The packed polyominoes must not overlap with each other.
The optimization objective is to pack all the polyominoes into the fewest canvases possible without any polyominoes overlapping.

\subsubsection{FFD Polyomino Packing Algorithm}
\label{sec:2d-ffd-polyominoes-packing}
Solving the packing problem is NP-hard (proof in \autoref{sec:np-packing}).
To obtain an approximate solution,
we designed an
algorithm inspired by the first-fit-descending (FFD) bin-packing~\cite{baker1985ffd}.
The algorithm sorts polyominoes by descending size
and places each polyomino into the first existing canvas where it fits without overlap.
If no existing canvas can fit it,
the algorithm opens a new canvas.
Unlike one-dimensional bin packing,
packing feasibility depends on both the remaining free area and the geometric placement of the polyomino.
This heuristic does not guarantee an optimal packing,
but it keeps the
packing
efficient while
targeting the objective of minimizing the number of canvases.

To describe the algorithm,
we use the following notation.
$P$
is
the list of
polyominoes resulting from the pruning process in \autoref{sec:prune};
$\mathcal{C}$
is
the set of canvases currently available for packing; and 
$h$ and $w$
are
the canvas height and width in tiles.
The algorithm \revAuthor{follows}:%
\begin{lstlisting}[language=Python, caption=FFD Polyomino Packing Algorithm,
    label=lst:packing-algorithm, basicstyle=\ttfamily\footnotesize,
    escapeinside={(*@}{@*)}]
def tryPlace(p: Polyomino, c: Canvas):
    for i in range(c.height - p.height + 1):
        for j in range(c.width - p.width + 1):
            if canPlaceAt(p, c, (i, j)): return i, j (*@\label{lst:can-place-at}@*)
def pack(P: list[Polyomino], h: int, w: int): (*@\label{lst:pack-def}@*)
    C: set[Canvas] = {} (*@\label{lst:canvas-init}@*)
    for p in sorted(P, key=len, reverse=True): (*@\label{lst:polyomino-sort}@*)
        position = None
        for c in C: (*@\label{lst:canvas-iter}@*)
            if position := tryPlace(p, c): break (*@\label{lst:try-place}@*)
        if position is None: C |= {c := newCanvas(h, w)} (*@\label{lst:new-canvas}@*)
        markPlacement(p, c, position or (0, 0)) (*@\label{lst:mark-placement}@*)
        yield p, c, position or (0, 0) (*@\label{lst:yield-placement}@*)
\end{lstlisting}

{\tt pack} (\autoref{lst:pack-def}) first sorts the polyominoes by size in descending order (\autoref{lst:polyomino-sort}).
Then, for each polyomino, it tries to place the polyomino into the first canvas that can fit the polyomino (\autoref{lst:canvas-iter}).
{\tt tryPlace} (\autoref{lst:try-place}) tries to place the polyomino into the canvas.
It returns the first position where the polyomino can be placed,
or {\tt None} if the polyomino cannot be placed in the canvas.
If the polyomino cannot be placed into any canvas,
a new canvas is created and the polyomino is placed at $(0, 0)$ in the new canvas (\autoref{lst:new-canvas}).
The overall runtime is $O(n \cdot (\log n + m \cdot h \cdot w \cdot |p|))$,
where $n$ is the number of polyominoes, $m$ the number of canvases, and $|p|$ the largest polyomino size.
After the placements are determined,
\sys{} renders the pixel content of the polyominoes into the canvases.

\subsection{Detect, Unpack, and Track}
\label{sec:detect-unpack-track}

Recall from \autoref{sec:tracking-by-detection} that $N$ is the number of input frames
and $N'$ is the number of object detector calls;
the reference pipeline issues $N' = N$ calls,
one per frame.
\sys{}'s preceding stages,
relevance classification,
pruning,
and packing,
aggregate the surviving polyominoes across frames into a set of canvases $\mathcal{C}$.
Because \sys{} invokes the detector once per canvas,
$N' = |\mathcal{C}| \le N$.
The \revAuthor{3} stages therefore reduce the number of detection calls from $N' = N$ in the reference pipeline to $N' = |\mathcal{C}|$ in \sys{},
which can be substantially smaller.
We now describe the remaining steps:
running the detector on each packed canvas,
unpacking the resulting bounding boxes to original frame coordinates,
and associating detections into tracks.

\subsubsection{Detection in Canvases}
\sys{} passes each packed canvas to the user-provided object detector.
Because each canvas has the same resolution as the original video frame,
objects appear at the same scale as in the reference pipeline,
\revRone{as the detector expects.}

A single detector call on a canvas is also sufficient to recover
the detections for every polyomino packed into it.
The total number of detector calls is therefore reduced from $N$
in the reference pipeline to $|\mathcal{C}|$,
which is the quantity minimized by \autoref{sec:relevance-classification}, \autoref{sec:prune}, and \autoref{sec:pack}.

\subsubsection{Unpacking Object Bounding Boxes}
The detector outputs bounding boxes in canvas coordinates,
which \sys{} maps back to the original frame coordinates.
\revAuthor{For each box, \sys{} identifies the polyomino containing its center within the canvas,
and subtracts the polyomino's placement offset to recover its original position.}

\subsubsection{Tracking}
\label{sec:track}
The unpacked detections are in the original per-frame coordinate system,
so \sys{} feeds them to the tracker unchanged,
producing tracks in the original video coordinate system.

\section{Learned Artifacts and Pareto Frontier}
\label{sec:e2e}

Before extracting tracks on the full video dataset via the \sys{} execution engine (\autoref{sec:exec}),
\sys{} learns dataset-specific artifacts and generates a Pareto frontier of system configurations.
This is done once per dataset and using small training and validation samples.
\sys{} chooses the training and validation samples at random;
their sizes are user-settable,
and our evaluation uses at most 60 one-minute videos per sample (\autoref{sec:evaluation}).
They produce the learned artifacts and a throughput-accuracy frontier,
not the final tracks.

First,
\sys{} trains a relevance classifier specialized to the user-provided detector,
which the execution engine uses to identify relevant tiles (\autoref{sec:specialization}).
Second,
\revRnine{for a given tracker,}
it measures empirical per-tile mistrack rates
\revRnine{to derive
$\bar{G}^{\bar{M}}$
and passes
the resulting matrix
$\bar{G}^{\bar{M}}$}
to the ILP for tile-granularity pruning (\autoref{sec:learn-constraints}).
Third,
it generates a Pareto-optimal throughput-accuracy frontier by sweeping system configurations (\autoref{sec:pareto-frontier}),
from which the user selects an operating point.
After these steps,
the \sys{} execution engine (\autoref{sec:exec}) extracts tracks on the remaining data using the selected configuration.

The second phase is the most technically challenging because we must choose tile sampling gaps without exhaustively searching all assignments.
A tile sampling gap specifies how many frames may elapse before that tile must be processed again.
Because different parts of a stationary scene have different motion patterns,
the best gap may vary across the $H \times W$ tile grid.
Exhaustively choosing one gap from $\Gamma$ for every tile would require evaluating $|\Gamma|^{H \times W}$ assignments,
which is computationally infeasible.
\sys{} therefore learns these gap choices through a scalar mistrack-rate tolerance $\bar{M} \in [0, 1]$.
For each tile $(i, j)$ and each candidate gap $\gamma \in \Gamma$,
\sys{} measures the empirical mistrack rate on the training sample.
Given a tolerance $\bar{M}$,
\sys{} chooses the largest gap whose measured mistrack rate is at most $\bar{M}$ for each tile.
The resulting per-tile choices form the maximum-gap matrix $\bar{G}^{\bar{M}}$.
Sweeping $\bar{M}$ produces a sequence of candidate matrices,
turning the exponential gap-assignment problem into a one-dimensional sweep.
Each candidate matrix is still evaluated end-to-end by running the \sys{} execution engine on a held-out validation sample.

\subsection{Training the Relevance Classifier}
\label{sec:specialization}
\sys{} trains the relevance classifier to imitate the user-supplied object detector at the tile level,
adapting the model-specialization paradigm~\cite{kang2017noscope}
from frame-level to tile-level granularity.
It runs the detector on the training video samples
and labels each tile as relevant if any detector bounding box overlaps its region.
\sys{} trains a lightweight relevance classifier by minimizing binary cross-entropy between the predicted relevance score $Score_{f,i,j}$
(\autoref{eq:classify})
and the detector-derived label.
This learned classifier is the first learned artifact, used by the execution engine
in \autoref{sec:relevance-classification}
(shown in \autoref{fig:system} \panellabelcolor{red}{A.1}).

\subsection{Learning Maximum Tile Sampling Gaps}
\label{sec:learn-constraints}

The second learned artifact is $\bar{G}^{\bar{M}}$,
a matrix of maximum tile sampling gaps that lets the execution engine prune polyominoes at tile granularity.
As discussed at the beginning of \autoref{sec:e2e},
learning $\bar{G}^{\bar{M}}$ turns an exponential joint search over per-tile gaps into a one-dimensional sweep over the mistrack-rate tolerance~$\bar{M}$.
For each tile $(i, j)$ and candidate sampling gap $\gamma \in \Gamma$,
\sys{} measures an empirical mistrack rate from the training sample.
Given a chosen tolerance~$\bar{M}$,
\sys{} derives $\bar{G}^{\bar{M}}$ by selecting gaps from the measured rates under this tolerance, which the ILP (\autoref{sec:prune}) adheres to at runtime,
as illustrated in \autoref{fig:system} \panellabelcolor{red}{A.2}.

Instead of enumerating the exponential space of $|\Gamma|^{H \times W}$ gap assignments,
we parameterize the selection through a single scalar mistrack-rate tolerance~$\bar{M}$.
At each tile, we take the largest sampling gap whose empirical mistrack rate does not exceed~$\bar{M}$.
In general, tiles with higher mistrack rates are assigned smaller maximum gaps,
while tiles with lower mistrack rates receive larger maximum gaps.
Varying~$\bar{M}$ deterministically yields different learned gaps.
Specifically, the measured rates define a deterministic mapping $f: [0, 1] \to \Gamma^{H \times W}$ where $f(\bar{M}) = \bar{G}^{\bar{M}}$.
This collapses the combinatorial per-tile search into a one-dimensional search over~$\bar{M}$,
performed when generating the Pareto frontier (\autoref{sec:pareto-frontier}).
In this sense,
sweeping~$\bar{M}$ approximates an exhaustive sweep over
$\bar{G} \in \Gamma^{H \times W}$.

\begin{figure}[t]
    \centering
    \resizebox{\columnwidth}{!}{\input{figures/generated/missrate_to_g_mapping.tex}}
    \includegraphics[width=.75\columnwidth]{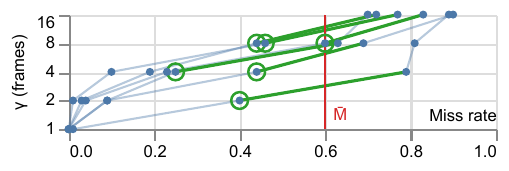}

    \Description{Two-by-three matrices of measured per-tile mistrack rates for sampling gaps gamma=2, gamma=4, and gamma=8 on top,
    connected by a downward arrow labeled with the gap-selection rule to two-by-three matrices of derived maximum tile sampling gaps for tolerances M-bar=0.4, M-bar=0.6, and M-bar=0.8 on the bottom.
    Below the matrices, a line chart plots miss rate on the horizontal axis and gamma in frames on a logarithmic vertical axis.
    Each line connects the five measured candidate gaps for one tile in the crop.
    This example rounds fractional miss rates to two decimal places before selecting gaps.
    A red vertical line marks M-bar=0.6; green segments cross it, and their circled left endpoints show the selected gaps.}
    \caption{Mapping from measured mistrack rates to maximum tile sampling gaps on a
    \revRone{$2 \times 3$} tile crop from B3D4.
    \revRone{
    In the line chart,
    each line represents one tile position,
    connecting its measured miss rates at $\gamma\in\{1,2,4,8,16\}$ frames.}
    \revRone{For each tile,
    we select the largest
    candidate
    gap whose miss rate is at most~$\bar M$.}
    \revRone{At $\bar M=0.6$ tolerance boundary,
    each tile's point on or immediately left of it indicates its selected gap (circled).}
    }
    \label{fig:missrate-to-g}

\end{figure}

Given~$\bar{M}$,
\sys{} derives $\bar{G}^{\bar{M}}$ by applying the gap-selection rule to the measured rates,
as illustrated in \autoref{fig:missrate-to-g}.
The resulting matrix has size $H \times W$
and the ILP (\autoref{sec:prune}) strictly adheres to these maximum tile sampling gaps, defined as
\[
\bar{g}^{\bar{M}}_{i,j} = \max
\bigl\{ \gamma \in \Gamma \mid \mathrm{MissRate}_{i,j}^{(\gamma)} \leq \bar{M} \bigr\},
\]
with $\bar{g}^{\bar{M}}_{i,j}$ defaulting to~$1$ if no gap in~$\Gamma$ satisfies the criterion.

\sys{} measures empirical mistrack rates as follows.
First,
we run the reference pipeline (\autoref{sec:tracking-by-detection}) on the training video samples
to obtain reference tracking results.
It then re-runs the tracker at each candidate sampling gap
in the set $\Gamma = \{1, 2, 4, 8, 16\}$,
where gap~$\gamma$ means the tracker processes every $\gamma$-th frame.
For each tile position $(i, j)$ and sampling gap $\gamma$,
\sys{} counts the number of tracking associations that differ
from the native-rate reference
and computes a per-tile empirical mistrack rate with:
\[
\mathrm{MissRate}_{i,j}^{(\gamma)}
= {(\mathrm{MissedA}_{i,j}^{(\gamma)} + 1)} / {(\mathrm{TotalA}_{i,j}^{(\gamma)} + 2)}.
\]
\revAuthor{When sampling every $\gamma$ frames,}
$\mathrm{MissedA}_{i,j}^{(\gamma)}$ is the number of incorrect associations
at tile $(i, j)$,
\revAuthor{and $\mathrm{TotalA}_{i,j}^{(\gamma)}$ counts evaluated associations at tile $(i,j)$.}
Laplace smoothing avoids \revAuthor{division by zero.}

Computing $\bar{g}^{\bar{M}}_{i,j}$ as $\max\{\gamma \in \Gamma \mid \mathrm{MissRate}_{i,j}^{(\gamma)} \leq \bar{M}\}$ tile-by-tile
is a heuristic approximation of the combinatorial per-tile search,
as it assumes per-tile mistrack rates are independent of other tiles' gaps.
We empirically show in \autoref{sec:ablation-constraints}
that $\bar{G}^{\bar{M}}$ lies on or near the Pareto frontier
of an exhaustive sweep of all per-tile gap assignments.

\subsection{Generating the Pareto Frontier}
\label{sec:pareto-frontier}

The Pareto frontier is computed using the validation sample
and produces a throughput-accuracy curve, as shown in \autoref{fig:system} \panellabelcolor{violet}{B}.
This curve is a user-visible tuning aid rather than the final tracking output.
Each point on the curve corresponds to a system parameter configuration,
recording the runtime throughput and tracking accuracy achieved under that configuration.
The user selects an operating point from this curve
based on a throughput constraint or an accuracy constraint, and \sys{} materializes the tracks on the rest of the input videos.

\sys{} first executes the reference pipeline (\autoref{sec:tracking-by-detection})
on the validation video samples to obtain reference tracking results.
Because the input video lacks ground-truth tracks,
these results serve as the tracking references
against which
\sys{} measures the accuracy of all optimized configurations.

\textbf{\em Additional Parameters.}
\label{sec:additional-knobs}
In addition to~$\bar{M}$, $T_r$, and $P$,
\sys{} sweeps the whole-frame sampling rate~$s$ and the tracker choice~$A$.
The sampling rate~$s$,
described in \autoref{sec:exec},
retains every $s$-th frame before the three \sys{} operators run.
Such whole-frame sampling is a common practice in video analytics systems~\cite{bastani2022otif,zhang2025hippo}.
The tracker choice~$A$ determines which tracker the execution engine invokes (\autoref{sec:track}).
\sys{}'s data model and execution engine are tracker-agnostic.
The sweep considers the user-provided tracker and SORT~\cite{bewley2016sort},
a lightweight Kalman-filter~\cite{kalman1960kalman} tracker that \sys{} uses as a built-in default.
Because SORT operates only on bounding-box geometry,
\sys{} includes it without video-specific retraining.
For each tracker option,
\sys{} repeats the measurement procedure of \autoref{sec:learn-constraints}.
Given a tolerance~$\bar{M}$,
\sys{} derives the corresponding
$\bar{G}^{\bar{M}}$ for the current tracker option and passes them to the ILP;
the configuration sweep treats~$A$ as a discrete parameter,
and each $(\bar{M}, A)$ pair yields a distinct operating point.
Additional trackers can be added at the cost of additional measurement and sweep time.

\textbf{\em Configuration Sweep.}
\sys{} then extracts tracks via its execution engine (\autoref{sec:exec})
over a sweep of parameter configurations (shown in \autoref{tab:frontier-sweep}),
recording each configuration's runtime throughput
and its tracking accuracy (HOTA~\cite{luiten2021hota}) against the reference tracking results.
This step uses the measured empirical rates from \autoref{sec:learn-constraints}
to instantiate $\bar{G}^{\bar{M}}$ for each tolerance value~$\bar{M}$.
Rather than sweeping maximum tile sampling gaps independently,
which would require enumerating $\Gamma^{H \times W}$ possible maximum-gap matrices,
the configuration sweep varies only the mistrack-rate tolerance~$\bar{M}$.
The mapping learned in \autoref{sec:learn-constraints} converts each tolerance into a full matrix~$\bar{G}^{\bar{M}}$,
so sweeping~$\bar{M}$ acts as a tractable approximation to sweeping the exponential space of maximum-gap matrices.
We employ a straightforward configuration sweep over the remaining parameter space%
,
and \sys{}'s sweep covers the five parameters in \autoref{tab:frontier-sweep}.

\textbf{\em Pareto Frontier.}
From all swept configurations,
\sys{} extracts the Pareto-optimal subset.
These are configurations for which no other configuration achieves both higher throughput and higher accuracy.
These Pareto-optimal points form the throughput-accuracy frontier.
An example is shown by the red curve in \autoref{fig:system} \panellabelcolor{violet}{B}.
Users can specify a throughput constraint and \sys{} selects the most accurate configuration meeting it,
or an accuracy constraint and \sys{} selects the highest-throughput configuration meeting it,
or pick a configuration directly from the frontier.

Once the user selects an operating point,
the chosen configuration and the learned artifacts are passed to
the \sys{} execution engine (\autoref{sec:exec}),
which extracts tracks on the remaining video data.
\sys{} retains every $s$-th frame from the input video
and discards the rest.
On each retained frame,
the specialized relevance classifier (\autoref{sec:specialization})
scores tiles against threshold~$T_r$.
The polyomino-pruning ILP (\autoref{sec:prune})
enforces the maximum tile sampling gaps~$\bar{G}^{\bar{M}}$
learned in \autoref{sec:learn-constraints},
derived from the chosen tolerance~$\bar{M}$.
Surviving polyominoes are padded with the pattern~$P$
and packed into canvases (\autoref{sec:pack}).
Finally, detection runs on the canvases
and the selected tracker~$A$ produces the tracks for the full dataset,
as shown in \autoref{fig:system} \panellabelcolor{orange}{C}.

\section{Evaluation}
\label{sec:evaluation}

We evaluate \sys's efficiency in tracking objects while maintaining tracking fidelity, comparing against prior systems on seven datasets.

\begin{figure*}[ht]
    \centering
    \includegraphics[width=\textwidth]{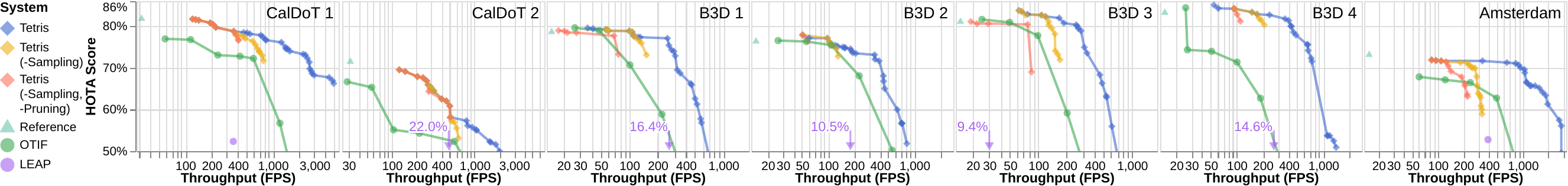}

    \caption{\revRone{End-to-end
    tracking accuracy versus throughput for \sys{}, the reference pipeline, prior systems, and ablations.}}

    \Description{Seven panels, one per dataset (CalDoT1, CalDoT2, B3D1 to B3D4, and Amsterdam), plot HOTA score from 50\% to 86\% against extraction throughput in frames per second on a logarithmic axis.
Each panel shows Pareto curves for Tetris, Tetris without whole-frame sampling, and Tetris without sampling and pruning, an OTIF curve, a single reference-pipeline point, and a single LEAP point.
Where LEAP's HOTA falls below the plotted range, a labeled downward arrow marks its throughput and gives its HOTA.}
    \label{fig:e2e-results}
\end{figure*}

\textbf{\em Accuracy Metrics.}
Prior systems evaluate accuracy using predefined queries~\cite{kang2017noscope,bastani2022otif,xu2024leap},
which can restrict evaluation to anticipated situations and fail to capture more complex object behaviors.
We instead adopt HOTA (\autoref{sec:track-materialization}) as a query-agnostic measure of track reconstruction quality.
HOTA
balances detection accuracy with trajectory association accuracy,
so missing detections,
fragmented tracks,
and merged identities all reduce the score.

\textbf{\em Datasets.}
We evaluate \sys{} on seven video datasets (CalDoT1/2, B3D1/2/3/4, and Amsterdam) that vary in scene density, camera viewpoint, and object motion patterns,
allowing us to test performance across a diverse set of real-world conditions.
CalDoT1/2 (720 $\times$ 480 pixels) were introduced by OTIF~\cite{bastani2022otif},
while Amsterdam (1280 $\times$ 720 pixels) was introduced by NoScope~\cite{kang2017noscope}.
We add the B3D (1080 $\times$ 720 pixels) datasets
(B3D1/2/3/4)
to expand the evaluation beyond existing benchmarks.
All datasets consist of approximately one-minute clips captured at 15 FPS;
CalDoT1/2,
and Amsterdam each include 180 clips,
while each B3D dataset includes 18 clips.

CalDoT1/2~\cite{bastani2022otif} consist of highway traffic footage from California Department of Transportation cameras;
Amsterdam~\cite{kang2017noscope} is captured from an urban street-side camera.
These datasets exhibit structured motion in relatively constrained settings.
The B3D datasets are drawn from the Berkeley DeepDrive Drone Dataset~\cite{wu2025b3d},
consisting of aerial footage of unsignalized intersections and roundabouts with less predictable motion and higher interaction density.
We mask non-road and parking areas in the B3D videos.
By including B3D, we extend beyond the structured highway and low-density roadway settings of CalDoT and Amsterdam.

For CalDoT1/2 and Amsterdam,
we partition the data into training,
validation,
and testing splits,
each consisting of 60 one-minute clips.
For the B3D datasets,
because of the amount of footage available,
we use training,
validation,
and testing splits of 8,
5,
and 5 approximately one-minute clips,
respectively.

\textbf{\em Prior Systems.}
We compare \sys{} against two prior approaches: OTIF and LEAP.
OTIF utilizes a lightweight proxy model to identify irrelevant regions within frames for more efficient batching, along with uniform frame sampling and a specialized tracker to further reduce calls to the object detector.
LEAP uses predictive sampling to prune frames that are unlikely to contain relevant information.
For both methods,
we use the authors'
available
implementations.

For OTIF,
we use the training split for its recurrent rate tracker,
since the original ``tracker'' split is unavailable across all datasets.

For a fair comparison across systems,
we standardize the object detector used in all experiments.
Specifically,
we use YOLOv5~\cite{khanam2024yolov5} for the CalDoT1, CalDoT2, and Amsterdam datasets,
and RetinaNet~\cite{lin2017focal} for the B3D datasets,
i.e., the detector released with B3D.
This ensures that any throughput and accuracy differences arise from the system rather than underlying detection models.
For \sys{},
the user-provided tracker is BYTETrack and the built-in default is SORT,
so Pareto frontier generation searches over the two tracker options $\{\text{BYTETrack}, \text{SORT}\}$.
When generating the Pareto frontier (\autoref{sec:pareto-frontier}),
BYTETrack~\cite{zhang2022bytetrack} at full frame rate serves as the user-provided tracker in the
reference pipeline~(\autoref{sec:tracking-by-detection}), and
the Pareto- optimal operating points may draw from either tracker option.

\textbf{\em Ground Truth.}
We obtain ground truth detection and tracking annotations for the test splits used in our end-to-end evaluation (\autoref{sec:e2e-results}).
For CalDoT1 and CalDoT2,
we use the dense per-frame object detection labels provided by the OTIF dataset.
For Amsterdam,
we start from the same OTIF dense labels,
filter to vehicle classes only,
and apply minor manual corrections to remove erroneous annotations.
For B3D1/2/3/4,
\revRone{we generate dense detection labels using
Faster R-CNN~\cite{ren2016fasterrcnn} trained with training labels from the B3D paper.%
}
Ground truth tracking annotations for all datasets are generated using OC-SORT~\cite{cao2023ocsort}%
.
\revRone{%
The B3D automated detection and tracking labels are verified and corrected by the authors.}

OC-SORT is used only for ground truth generation,
ensuring that evaluation results are not biased by shared trackers.
\sys does not
\revRone{have access to the ground-truth labels}
while learning the artifacts or generating the Pareto frontier.

\textbf{\em Detector Training.}
We train the YOLOv5 models on the training splits using the same curated labels described above:
the dense detection labels from the OTIF dataset for CalDoT1 and CalDoT2,
and the vehicle-filtered,
manually corrected OTIF labels for Amsterdam.

\textbf{\em Experiment Setup.}
All evaluated systems consist of two stages:
a preparation stage to learn from the training split and generate their configurations using the validation split,
and a track extraction stage to track objects in the test split.
We report throughput on track extraction only.
\revRone{For \sys{},
this includes relevance classification,
polyomino grouping,
ILP construction and solving,
packing,
detection,
unpacking,
tracking,
and interpolation.}
Each data point in the results corresponds to a distinct system configuration.
For \sys{} and OTIF,
configurations are the Pareto-optimal operating points selected on the validation set (\autoref{sec:pareto-frontier}).
LEAP, having a single fixed configuration, contributes one point.
For \sys{} and OTIF,
we report only the Pareto-optimal points from the test split results.

\textbf{\em Hardware.}
All experiments are conducted on a server equipped with dual Intel Xeon Gold 6248 CPUs
(2.50\,GHz),
504\,GB of RAM,
and an NVIDIA Quadro RTX 6000 GPU
(24\,GB VRAM).
Reported runtimes measure computation time only,
excluding disk I/O.

\textbf{\em Track Interpolation.}
Tracking outputs from all systems are linearly interpolated to produce detection points on every frame for every track.
This step ensures a fair comparison under the HOTA metric,
which penalizes missing per-frame detections within tracks.
Systems that employ frame skipping inherently produce tracks with detection gaps;
without interpolation,
these gaps would exaggerate the HOTA penalty even when the underlying motion is faithfully captured,
for example,
a vehicle traveling at constant speed along a straight path.
By applying interpolation uniformly across all systems,
we isolate tracking quality from the detection gaps that frame-sampling strategies introduce.

\subsection{End-to-end Results}
\label{sec:e2e-results}

\autoref{fig:e2e-results} summarizes the throughput-accuracy tradeoff across all seven datasets.
\sys{} consistently dominates the Pareto frontier,
achieving higher HOTA scores than both OTIF and LEAP at the same throughput,
and higher throughput at the same accuracy level.
At matched throughput,
we compare each prior work's Pareto-optimal configuration against the most accurate \sys{} configuration that is at least as fast.
\revAuthor{\sys{} achieves higher HOTA in \compareMatchedTputHigherCount{} of these \compareMatchedTputReachableCount{} comparisons
(out of \compareMatchedTputPriorConfigCount{} prior works' configurations).
In the remaining comparison,
on \compareMatchedTputNotHigherDatasets{},
the prior configuration is ahead by \compareMatchedTputNotHigherMaxGap{} HOTA.}
Of the remaining \compareMatchedTputUnreachableCount{} prior configurations that exceed \sys{}'s highest throughput,
\compareMatchedTputUnreachableZeroHotaCount{} produce zero HOTA;
the last achieves a \compareMatchedTputUnreachableRestSpeedup{}$\times$ speedup over \sys{}'s fastest configuration
while dropping HOTA by \compareMatchedTputUnreachableRestHotaDrop{}
on the \compareMatchedTputUnreachableRestDataset{} dataset.
The median HOTA gain across these comparisons is \compareMatchedTputMedianHotaGain{},
per-dataset maximum gains range from \compareMatchedTputPerDatasetMaxGainMin{} to \compareMaxHotaImprovement{},
and the largest gain occurs at \compareMaxHotaImprovementFps{} FPS on the \compareMaxHotaImprovementDataset{} dataset.

In realistic deployment scenarios where tracking accuracy is critical,
we measure the maximum speedup each system achieves over the full-frame,
every-frame reference pipeline at bounded HOTA accuracy loss.
\sys{} achieves a speedup while staying within a 5\% HOTA drop from the reference pipeline on all
\revisionDatasetCount{} datasets,
while prior systems meet this constraint on only \comparePriorMeetFivePct{}/\revisionDatasetCount{} datasets.
To avoid overclaiming \sys{}'s improvement over prior systems,
we restrict prior systems to Pareto-optimal configurations and compare against a prior configuration whose HOTA is no higher than \sys{}'s selected configuration.
Under this 5\% constraint,
\sys{} achieves \compareAccuracyMatchedSpeedupMinFivePct{}$\times$ to \compareAccuracyMatchedSpeedupMaxFivePct{}$\times$ speedup over prior systems,
and \compareNaiveSpeedupMinFivePct{}$\times$ to \compareNaiveSpeedupMaxFivePct{}$\times$ speedup over the reference pipeline.
At a 10\% HOTA drop,
\sys{} achieves \compareAccuracyMatchedSpeedupMinTenPct{}$\times$ to \compareAccuracyMatchedSpeedupMaxTenPct{}$\times$ speedup over prior systems,
and \compareNaiveSpeedupMinTenPct{}$\times$ to \compareNaiveSpeedupMaxTenPct{}$\times$ speedup over the reference pipeline.
Under both constraints,
\sys{} consistently achieves higher speedup than all other systems on every dataset.

\textbf{\em Analysis.}
LEAP is designed to optimize for object-existence queries,
such as counting or detecting the time period during which an object is present,
rather than reconstructing full per-frame object tracks.
Accordingly,
LEAP outputs a time interval and \revRone{an assigned} motion pattern for each detected object rather than a per-frame trajectory.
\revRone{To evaluate LEAP with HOTA,
we reconstruct per-frame trajectories from LEAP's sampled detection boxes and assigned motion patterns
through 3 approaches.
(i) We keep LEAP's detected boxes and linearly interpolate between consecutive boxes of the same track.
(ii) We use LEAP's assigned motion pattern and resample it to produce one box per frame over its predicted presence interval.
(iii) We keep the same detected boxes as in (i) and align the assigned motion pattern to them,
matching each box's center to the closest point on the pattern;
we then interpolate along the aligned pattern between consecutive detections.
LEAP learns its motion patterns from the training clips during preparation.
All three methods preserve LEAP's track IDs.
The end-to-end comparison uses the highest HOTA of the three reconstructions for each dataset.}

\revRone{%
LEAP achieves its highest HOTA on CalDoT1 and Amsterdam,
where small position changes between consecutive sampled
frames
aid reconstruction from sampled detections.}
\revRone{On CalDoT2, LEAP's median sampling interval is \revisionLeapCalDoTTwoMedianSamplingSeconds{} seconds.
Across all ground-truth vehicle tracks,
\revisionLeapCalDoTTwoFewerThanTwoSamplesPct{}\% appear in fewer than 2 frames sampled by LEAP,
providing insufficient detected bounding boxes for interpolation.}
\revRone{B3D captures unsignalized intersections and traffic circles,
where varied routes and vehicle visibility durations can limit motion-pattern prediction and predictive sampling.
Vehicles stay visible for \revisionLeapBThreeDVisibilityTenSeconds{}--\revisionLeapBThreeDVisibilityNinetySeconds{}\,s within a clip (10th--90th percentile).}

On the other hand, OTIF achieves reasonable accuracy,
but its speedup derives primarily from uniform frame skipping,
with its segmentation proxy model contributing comparatively little to throughput gains,
as reported by Bastani et al.~\cite{bastani2022otif}.
This reliance on frame skipping produces coarse-grained tracks that degrade accuracy at higher throughput operating points,
consistent with the observation that uniform temporal sampling cannot accommodate spatially varying sampling requirements (\autoref{sec:motivation}, Obs.~2).
The limited benefit of OTIF's proxy is visible directly in the B3D4 Pareto front.
Its most accurate point disables the segmentation proxy entirely,
while the next point enables it and exhibits a sharp accuracy drop with little speedup gain.
This points to a limitation that motivates \sys{}'s design:
OTIF's rectangular crops are a poor fit for the irregular shape of real activity regions (\autoref{sec:motivation}, Obs.~3),
forcing the proxy to either over-include background (yielding little speedup) or
discard tiles with detections (yielding the observed accuracy drop).

\begin{figure}[t]
\centering
\includegraphics[width=0.66\columnwidth]{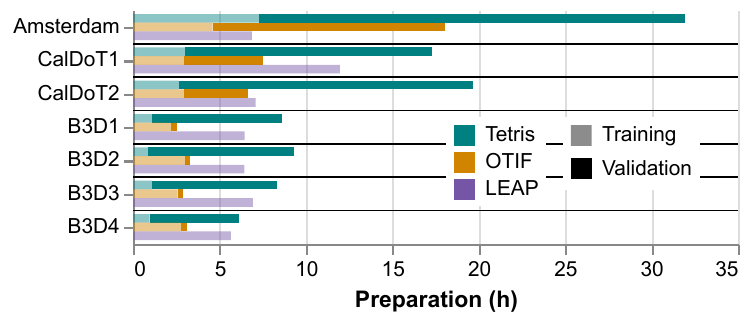}
\hfill
\includegraphics[width=0.30\columnwidth]{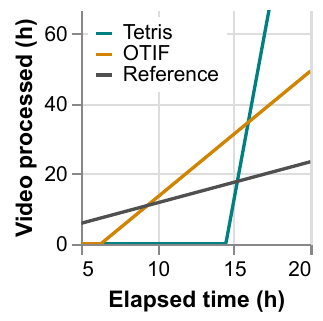}

\caption{\revRnine{\panellabel{Left} Preparation time split into training and validation.
\panellabel{Right} Video processed against elapsed time including preparation,
for \sys{}'s 5\% HOTA-loss configuration,
the paired OTIF configuration,
and the reference pipeline.%
}}
\Description{Training and validation preparation bars for Tetris, OTIF, and LEAP on all seven datasets, next to dataset-averaged progress curves for Tetris at 5 percent HOTA loss, paired OTIF configurations, and the reference pipeline.
Averaged across datasets.}
\label{fig:preparation-cost}\label{fig:amortization}

\end{figure}
\subsection{\revRnine{Preparation Cost and Amortization}}
\label{sec:preparation-cost}

\revRnine{\autoref{fig:preparation-cost} \panellabel{Left} compares
each system's preparation time
before extraction,
split into training and validation.
For \sys{},
training creates reference labels and learned artifacts (\autoref{sec:specialization}, \autoref{sec:learn-constraints});
validation generates the Pareto frontier by running and scoring all
configurations.
OTIF trains its proxy model and tracker,
then tunes configurations on validation videos.
LEAP trains its tracker model and learns motion patterns,
without a validation search.
Validation accounts for \revisionTetrisValidationShareMinPct{}--\revisionTetrisValidationShareMaxPct{}\% of \sys{}'s preparation time,
while classifier training takes only \revisionTetrisClassifierTrainingShareMinPct{}--\revisionTetrisClassifierTrainingShareMaxPct{}\%
and the detector pass over the training split takes \revisionTetrisTrainingDetectionShareMinPct{}--\revisionTetrisTrainingDetectionShareMaxPct{}\%.
In total,
\sys{} needs \revisionTetrisPreparationMinHours{}--\revisionTetrisPreparationMaxHours{} hours,
compared with \revisionOtifPreparationMinHours{}--\revisionOtifPreparationMaxHours{} hours for OTIF
and \revisionLeapPreparationMinHours{}--\revisionLeapPreparationMaxHours{} hours for LEAP.
As we will see,
the extra preparation time leads to substantial throughput gain.
}

\revRnine{\autoref{fig:amortization} \panellabel{Right} plots processed video over elapsed time,
averaged over datasets.
\revRnine{A system with preparation $P_h$ hours and throughput $F_s$ FPS processes $F_s\max(0,t-P_h)/15$ hours of 15-FPS video within $t$ hours.}
At the 5\% HOTA-loss configurations,
\sys{} overtakes the reference pipeline after \revisionCrossTetrisReferenceMinHours{}--\revisionCrossTetrisReferenceMaxHours{} hours.
At the matching 5\% loss configuration (\autoref{sec:e2e-results}),
\sys{} overtakes OTIF after \revisionCrossTetrisOtifMinHours{}--\revisionCrossTetrisOtifMaxHours{} hours.}

\revRnine{%
LEAP
only has 1 configuration,
and never reaches the 5\% bound on any dataset;
we separately compare it against \sys{}'s fastest configuration that is more accurate than LEAP.
\sys{} overtakes
LEAP after \revisionCrossLeapmatchedLeapMinHours{}--\revisionCrossLeapmatchedLeapMaxHours{} hours.}

\revRnine{\sys{} spends more preparation time to gain extraction throughput.
It recovers this cost within \revisionCrossTetrisLatestHours{}
elapsed hours
on every dataset,
after which it processes video \revisionSpeedupTetrisReferenceMin{}--\revisionSpeedupTetrisReferenceMax{}$\times$ faster than the reference pipeline
and \revisionSpeedupTetrisOtifMin{}--\revisionSpeedupTetrisOtifMax{}$\times$ faster than the paired OTIF configuration.}

\subsection{\revRone{Downstream Query Accuracy}}
\label{sec:leap-queries}

\revRone{%
To assess the downstream query accuracy derived from \sys{}'s materialized tracks,
we evaluate the frame-selection,
per-frame counting,
and top-10 queries defined by LEAP,
treating all vehicles as one class.
We compare against two kinds of prior systems.
First, OTIF and LEAP represent track-materializing approaches that answer queries from their tracks and presence intervals,
respectively.}
\revRseven{%
Second, to compare against query-specific approaches that output query results directly,
we evaluate NoScope~\cite{kang2017noscope} and Seiden~\cite{bang2023seiden}
on the queries they support
(NoScope supports only selection).
All systems
use the same reference detector as \sys{}.}

\revRseven{%
NoScope and Seiden trade accuracy for speed by adjusting their target error rates
and frame budgets,
respectively.}
\revRseven{We use NoScope's default target error rates
and set Seiden's frame budgets to 1/4, 1/8, 1/12, \ldots, 1/32 of the frames.}
\revRone{%
For each dataset and query,
we pick OTIF's, NoScope's, and Seiden's Pareto-optimal configurations closest to LEAP in FPS
and \sys{}'s slowest configuration that is faster than all of them.}
\revRone{Query scores (\%) in each cell below list \sys{} / OTIF / LEAP / Seiden / NoScope,
with the best value in bold:}
\begin{center}
\resizebox{.999\columnwidth}{!}{%
\begin{tabular}{@{}l*{3}{@{\quad}*{4}{r@{\hspace{2pt}}}r}@{}}
{\bfseries Dataset} & \multicolumn{5}{@{}r@{\quad}}{{\bfseries Selection (F1) $\uparrow$}} & \multicolumn{5}{@{}r@{\quad}}{{\bfseries Count (MAPE~\cite{makridakis1984mape}) $\downarrow$}} & \multicolumn{5}{@{}r@{}}{{\bfseries Top-10 (Precision) $\uparrow$}} \\
\arrayrulecolor{black}\midrule\arrayrulecolor{.}
CalDoT1 & 96.2 / & 96.8 / & 95.4 / & \textbf{98.0} / & 97.1 & 8.9 / & 10.0 / & 33.0 / & \textbf{7.3} / & -- & \textbf{100} / & \textbf{100} / & 90.0 / & \textbf{100} / & -- \\
CalDoT2 & 96.2 / & \textbf{97.5} / & 91.5 / & 96.8 / & 97.1 & 16.6 / & \textbf{15.4} / & 42.0 / & 22.6 / & -- & \textbf{90.0} / & \textbf{90.0} / & \textbf{90.0} / & \textbf{90.0} / & -- \\
B3D1 & \textbf{100} / & 99.9 / & \textbf{100} / & \textbf{100} / & 93.3 & 33.0 / & 38.7 / & 43.9 / & \textbf{11.3} / & -- & \textbf{90.0} / & \textbf{90.0} / & \textbf{90.0} / & \textbf{90.0} / & -- \\
B3D2 & \textbf{100} / & 99.9 / & \textbf{100} / & \textbf{100} / & \textbf{100} & 26.5 / & 28.9 / & 27.3 / & \textbf{17.5} / & -- & \textbf{70.0} / & \textbf{70.0} / & 50.0 / & \textbf{70.0} / & -- \\
B3D3 & \textbf{100} / & \textbf{100} / & \textbf{100} / & \textbf{100} / & \textbf{100} & 10.7 / & \textbf{5.9} / & 96.1 / & 6.0 / & -- & \textbf{90.0} / & \textbf{90.0} / & 80.0 / & \textbf{90.0} / & -- \\
B3D4 & \textbf{100} / & 99.0 / & \textbf{100} / & \textbf{100} / & 85.9 & 15.7 / & 39.5 / & 124 / & \textbf{11.1} / & -- & \textbf{90.0} / & \textbf{90.0} / & 60.0 / & \textbf{90.0} / & -- \\
Amsterdam & \textbf{98.3} / & 97.9 / & 98.2 / & 98.2 / & 94.5 & \textbf{18.3} / & 23.3 / & 59.3 / & 26.1 / & -- & \textbf{100} / & \textbf{100} / & 90.0 / & 90.0 / & -- \\
\end{tabular}
}
\end{center}

\revRone{%
Among track-materializing approaches,
\sys{} is the best in \revisionQueryBestPct{}\% of
cases,
within 1 percentage point in \revisionQueryComparablePct{}\%,
and at most \revisionQueryWithinFiveMaxGapPp{} points behind in the remaining \revisionQueryWithinFivePct{}\%.}
\revRseven{%
Across all systems,
\sys{} is the best in \revisionQueryAllBestPct{}\%,
at most \revisionQueryAllWithinFiveMaxGapPp{} points behind in \revisionQueryAllWithinFivePct{}\%,
\revisionQueryAllLowerSecondGapPp{} points behind in one case,
and \revisionQueryAllLowerMaxGapPp{} points behind in another.}
\revRseven{%
On B3D,
every test frame contains a vehicle,
so all systems except NoScope reach near-perfect F1,
which does so only at
\revisionNoScopeBThreeDStrictCostMin{}--\revisionNoScopeBThreeDStrictCostMax{}\% lower throughput.%
}

\revRone{These results support our claim, as \sys{} efficiently materializes tracks that accurately answer track-related queries,
while
remaining competitive
with systems specialized for individual queries.}

\subsection{Ablation Studies}
We now systematically break down the contribution of the operators in \sys{}'s tracking pipeline.
\revAuthor{We first evaluate the relevance classifier against an RGB-only baseline,
then measure} \revRone{how packing affects detector accuracy} \revAuthor{and how densely the packer fills the canvases.}
We examine the throughput improvements achieved by sequentially enabling our core spatial and temporal optimization operators,
and then whole-frame sampling.
We also validate the one-dimensional tolerance sweep against the full combinatorial per-tile gap search.
\revRnine{Finally,
we break down the extraction runtime by operator
and evaluate how the pruning ILP scales.}

\subsubsection{Relevance Classifier Accuracy}
\label{sec:ablation-classifier}

\begin{figure}[t]
    \centering
    \includegraphics[width=\columnwidth]{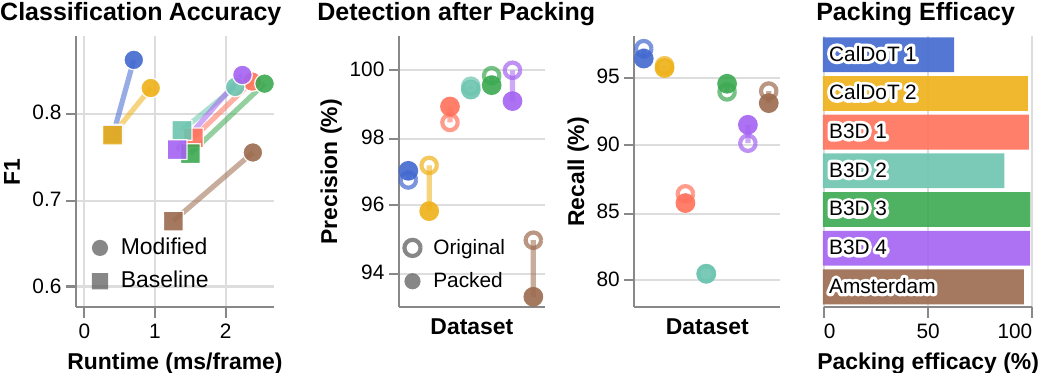}

    \caption{\revRnine{\panellabel{Left} Relevance-classifier comparison.}
    Modified:
    \autoref{sec:relevance-classification}.
    Baseline:
    RGB-only ShuffleNet~\cite{ma2018shufflenet}.
    \revRone{\panellabel{Middle} Detector precision and recall on original frames and packed canvases per dataset.%
    }
    \revRnine{\panellabel{Right} Packing efficacy and dataset color legend.}
    }
    \Description{Three parts.
Left: F1 against classifier runtime per frame for the modified and RGB-only baseline classifiers on each dataset, with a line joining each dataset's pair.
Middle: detector precision and recall per dataset, with hollow markers for original frames and filled markers for packed canvases, joined by a line.
Right: horizontal bars of packing efficacy per dataset, from 0 to 100 percent, labeled with the dataset names and colored as in the other panels.}
    \label{fig:classifier-and-packing}

\end{figure}

We compare the deployed relevance classifier against the original RGB-only ShuffleNet baseline.
Across \PtwozerofiveCompareClassifiersDatasetCount{} datasets at $T_r$=\PtwozerofiveCompareClassifiersThreshold,
adding the frame-difference image and position encoding improves F1 over the baseline on all \PtwozerofiveCompareClassifiersImprovedFOneDatasetCount{} datasets,
as shown in the \panellabel{Left} panel of \autoref{fig:classifier-and-packing}.
Our relevance classifier yields an average F1 gain of \PtwozerofiveCompareClassifiersMacroAvgFOneDeltaPp{}\%.
The largest gain appears on \PtwozerofiveCompareClassifiersBestDatasetDisplay{},
where F1 rises from \PtwozerofiveCompareClassifiersBestFOneBaseline{} to \PtwozerofiveCompareClassifiersBestFOneModified{}
(gain \PtwozerofiveCompareClassifiersBestFOneDeltaPp{}\%).
\revAuthor{Relative to full-frame detection,
classification time rises from \revisionRgbClassifierDetectorAvgPct{}\% for the RGB-only baseline to \revisionAugmentedClassifierDetectorAvgPct{}\% for our relevance classifier,
on average across datasets.}

\subsubsection{\revRone{Detection Accuracy after Packing}}
\label{sec:packed-detection}
\revRone{Packing places polyominoes from different frames and image locations side by side,
changing the detector's input distribution.
We quantify how packing affects detection accuracy on the test split.
To isolate its effect,
we select relevant tiles using ground-truth boxes and disable ILP-based pruning.
We sample test frames at $s$=16 and use the padding from each dataset's 5\% HOTA-loss configuration (\autoref{sec:e2e-results}),
with $P$=.5 on Amsterdam and CalDoT and $P$=0 on B3D.
We apply the reference detector to the original sampled frames and their packed canvases,
then map canvas detections back to source-frame coordinates.
We score both outputs against the ground truth,
including only annotations and detections whose centers lie in retained tiles.
We compute precision and recall using COCO's greedy box matching~\cite{cocoeval} with a
standard PASCAL VOC~\cite{everingham2010voc} minimum IoU of $.5$.
}

\revRone{With these padding settings,
packing changes precision and recall by at most \revisionPackingDeployedMaxChange{} percentage points (\panellabel{Middle} panels of \autoref{fig:classifier-and-packing}).
On Amsterdam and CalDoT,
packing without padding lowers recall by \revisionPackingSparseNoPadRecallDropMin{}--\revisionPackingSparseNoPadRecallDropMax{} points relative to detection on the original frames.%
}

\subsubsection{Packing Efficacy}
\label{sec:ablation-packing}
We quantify how tightly \sys{}'s packer fills each generated canvas by measuring the
fraction of total canvas tile-cells that the packer actually uses,
\revAuthor{$\text{packing efficacy} \;=\; \frac{\text{occupied tiles}}{\text{occupied tiles} + \text{empty tiles}}$},
where the denominator is
the number of canvases $\times$ canvas height $\times$ canvas width,
measured in tile units.
\revAuthor{We use classifier-selected tiles on the test split with $s$=1,
$T_r$=0.5,
$\bar{M}$=none,
and $P$=0.}
The packer utilizes
\packingEfficacyMeanAcrossDatasets{}\% of the available canvas area on average across
\packingEfficacyDatasetCount{} datasets,
peaking at \packingEfficacyBestValue{}\% on \packingEfficacyBestDatasetDisplay{}.
The \panellabel{Right} panel of \autoref{fig:classifier-and-packing} shows the per-dataset packing
efficacy.

The two least densely packed datasets are B3D~2 and CalDoT~1,
which fall noticeably below the
average
(down to \packingEfficacyWorstValue{}\%).
We traced both to dataset-specific properties of the relevant polyominoes.
B3D~2 captures a busy unsignalized intersection in which vehicles routinely cluster
shoulder-to-shoulder.
Adjacent tiles with vehicles merge into a single connected polyomino,
so the per-canvas polyomino population is shifted toward fewer,
larger polyominoes than on the less-congested B3D scenes.
Specifically, aggregated across all
\revAuthor{test}
videos,
B3D~2 produces \packingEfficacyBThreeDTwoPolyPerCanvas{} polyominoes per canvas
averaging \packingEfficacyBThreeDTwoPolyMeanTiles{} tiles each,
versus \packingEfficacyBThreeDThreePolyPerCanvas{} polyominoes per canvas
averaging \packingEfficacyBThreeDThreePolyMeanTiles{} tiles each on B3D~3.
CalDoT~1 exhibits a similar effect.
Aggregated across all
\revAuthor{test}
videos,
CalDoT~1 produces \packingEfficacyCalDoTOnePolyPerCanvas{} polyominoes per canvas
averaging \packingEfficacyCalDoTOnePolyMeanTiles{} tiles each,
versus \packingEfficacyCalDoTTwoPolyPerCanvas{} polyominoes per canvas
averaging \packingEfficacyCalDoTTwoPolyMeanTiles{} tiles each on CalDoT~2.

\subsubsection{Optimization Operators}
\label{sec:ablation-operators}
We measure the impact of each optimization technique by incrementally enabling them and comparing their throughput over the reference baseline under a strict 5\% maximum HOTA loss.
\sys{} first introduces the relevance classification and polyomino packing operators.
As shown by the Pareto front results,
this combined spatial pruning approach brings the throughput speedup over the reference execution up to \ablationOneSpeedupMax{}$\times$
(average \ablationOneSpeedupAvg{}$\times$)
while safely remaining within the tracking accuracy constraint.
Next, \sys{} adds the fine-grained polyomino pruning operator, which significantly accelerates tracking in simple, low-activity video regions.
The inclusion of polyomino pruning further increases the maximum speedup to \ablationTwoSpeedupMax{}$\times$
(average \ablationTwoSpeedupAvg{}$\times$)
over the baseline execution.
Finally, we evaluate enabling the whole-frame sampling knob $s > 1$ (\autoref{sec:additional-knobs}).
Together,
the complete execution engine achieves a maximum speedup of \ablationThreeSpeedupMax{}$\times$
(average \ablationThreeSpeedupAvg{}$\times$)
while still strictly adhering to the 5\% tracking accuracy loss constraint.

\subsubsection{Maximum tile sampling gaps}
\label{sec:ablation-constraints}

We evaluate the sweep over the mistrack-rate tolerance~$\bar{M}$ from \autoref{sec:learn-constraints}.
Each tolerance yields a different $\bar{G}^{\bar{M}}$.
We show that the sweep approximates the Pareto frontier of pruning ratio versus HOTA against obtained by exhaustively enumerating all possible $\bar{G}$.
To keep the exhaustive baseline feasible,
we reduce the tile grid to $3 \times 3$
and the candidate gap set to $\{1, 2, 4\}$,
yielding $3^9 \approx 20{,}000$ per-tile gap assignments.
We do not report throughput directly on this sweep,
because re-running full tracking execution on every assignment is prohibitively expensive;
instead,
we use the pruning ratio (fraction of tiles skipped)
as a throughput proxy.
We use BYTETrack as the tracker.

To isolate the effect of the maximum tile sampling gaps
from classifier and detector variance,
the experimental pipeline constructs polyominoes
from the reference-pipeline bounding boxes rather than classifier output,
retains only
boxes
inside a polyomino retained on the same frame,
and feeds
them
to BYTETrack.
We report HOTA against the
\revAuthor{ground-truth tracks}
alongside the pruning ratio.

\begin{figure}[t]
    \centering
    \includegraphics[width=\columnwidth]{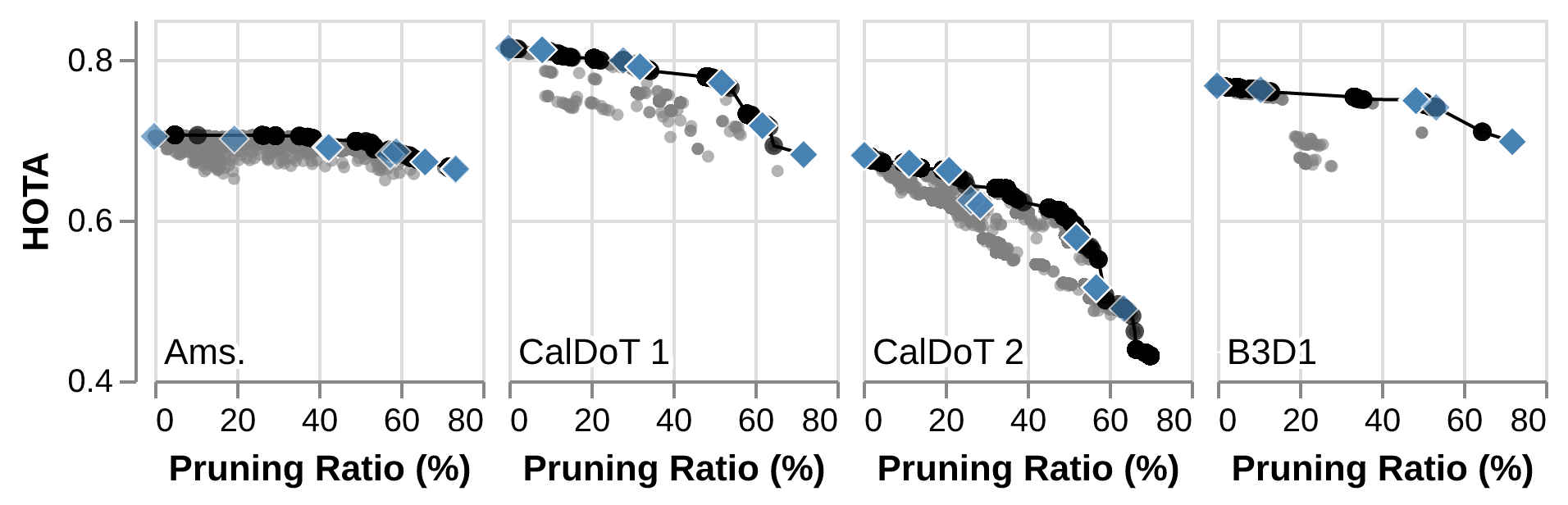}

    \caption{HOTA vs.\ pruning ratio per dataset.
    Gray: exhaustive per-tile sampling assignments
    over a $3\times 3$ grid with gaps $\{1, 2, 4\}$.
    Black: Pareto frontier of the exhaustive sweep.
    Blue diamonds: our heuristic $\bar{G}^{\bar{M}}$ swept over $\bar{M}$ with BYTETrack.}
    \Description{}
    \label{fig:sampling-constraints-ablation}

\end{figure}

\autoref{fig:sampling-constraints-ablation} shows the resulting trade-off per dataset.
The heuristic points (blue diamonds),
obtained by sweeping $\bar{M}$,
lie on or near the Pareto frontier of the exhaustive per-tile sweep across datasets,
indicating that the one-dimensional tolerance sweep loses little
relative to the full combinatorial search.
Quantitatively,
at each heuristic pruning level we compare against the exhaustive
Pareto point with the highest pruning ratio not exceeding that level;
HOTA loss versus that anchor is at most \MistrackHeuristicMaxHotaLossPercent\%{}
(mean \MistrackHeuristicAvgHotaLossPercent\%).
\subsubsection{\revRnine{Extraction Operator Costs Breakdown}}
\label{sec:ilp-cost}

\begin{figure}[t]
    \centering
    \includegraphics[width=\columnwidth]{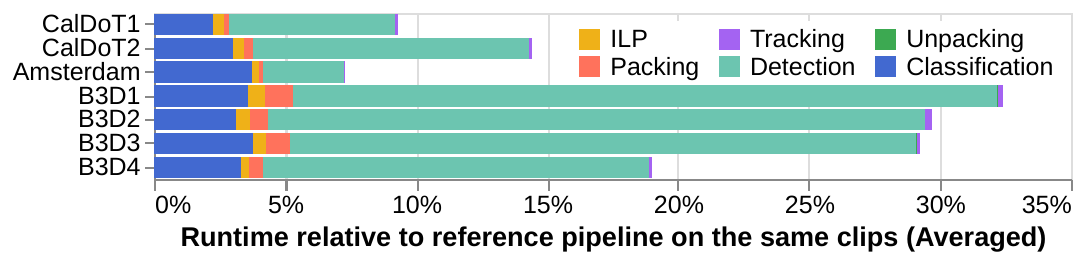}

    \caption{\revRnine{Runtime breakdown
    per operator.
    }}
    \label{fig:runtime-breakdown}
    \Description{One stacked bar per dataset for the fixed configuration, with operator runtimes expressed relative to the reference pipeline's total runtime.}

\end{figure}

\revRnine{\autoref{fig:runtime-breakdown} reports runtime breakdown for the
configuration 
(%
    $s$=1,
    $T_r$=0.5,
    $\bar{M}$=0.2,
    $P$=1,
    SORT%
)
across datasets%
.
Classification
takes
\revisionFixedClassifierPipelinePctMin{}--\revisionFixedClassifierPipelinePctMax{}\% of total extraction,
as it reduces downstream detection work.
Running classification on every frame, however, costs only \revisionFixedClassifierDetectorPctMin{}--\revisionFixedClassifierDetectorPctMax{}\% of all-frame detector compute time.
Both the classifier
and the detector process every pixel at native resolution,
so this overhead should stay similar for higher-resolution video.
Optimistically eliminating all other work gives a classifier-only speedup ceiling of \revisionFixedClassifierSpeedupCapMin{}--\revisionFixedClassifierSpeedupCapMax{}$\times$
at $s$=1
and \revisionFixedSampledSpeedupCapMin{}--\revisionFixedSampledSpeedupCapMax{}$\times$
at $s$=16.%
}
\revRnine{Pruning,
including ILP construction and solving,
takes \revisionFixedPruningPipelinePctMin{}--\revisionFixedPruningPipelinePctMax{}\% of extraction,
or \revisionFixedPruningNaivePctMin{}--\revisionFixedPruningNaivePctMax{}\% of reference-pipeline time.
Before presolve,
the largest ILP in each dataset has \revisionFixedIlpVariablesMin{}--\revisionFixedIlpVariablesMax{} variables and \revisionFixedIlpConstraintsMin{}--\revisionFixedIlpConstraintsMax{} constraints.}
\revRnine{Packing takes \revisionFixedPackingPipelinePctMin{}--\revisionFixedPackingPipelinePctMax{}\% of extraction,
or \revisionFixedPackingNaivePctMin{}--\revisionFixedPackingNaivePctMax{}\% of reference-pipeline time,
mostly canvas rendering (\revisionFixedRenderingPackingPctMin{}--\revisionFixedRenderingPackingPctMax{}\% of packing time).}

\subsubsection{\revRnine{ILP Scalability}}
\label{sec:ilp-scalability}

\revRnine{Since each ILP covers a one-minute window of video (\autoref{sec:prune}),
its size depends on the relevant tiles in that window rather than on video duration.
\sys{} limits each ILP to a \revisionFixedIlpBudgetSeconds{}-s solver budget.
Under the same
configuration as \autoref{sec:ilp-cost} and
across the \revisionFixedIlpRuns{} test clips,
\revisionFixedIlpInputTilesShort{} relevant tiles enter the ILP,
and pruning keeps \revisionFixedIlpPrunedTilesShort{} of them.
\revisionFixedIlpWithinPointOneSecond{} of the \revisionFixedIlpRuns{} ILPs finish within the budget.
To estimate the pruning this
budget leaves unrealized,
we rerun the same ILPs with a 1-hour budget.
In this rerun,
\revisionFixedIlpWithinOneSecond{} return within 1\,s,
\revisionFixedIlpWithinTenSeconds{} within 10\,s,
\revisionFixedIlpWithinSixtySeconds{} within 60\,s,
and \revisionFixedIlpWithinOneHour{} within 1\,h;
\revisionFixedIlpHitOneHour{} reach the cap.
Across all clips,
the 1-hour budget prunes an additional \revisionFixedIlpHourAllTilesReducedShort{} tiles out of the \revisionFixedIlpPrunedTilesShort{} tiles kept under the \revisionFixedIlpBudgetSeconds{}-s budget.
That is only \revisionFixedIlpHourAllTilesReducedPct{}\% of the tiles kept under the shorter budget.
The ILP's solve time therefore stays bounded per minute of video, with little pruning lost.}

\section{Conclusion}
\label{sec:conclusion}

We presented \sys{},
a track-extraction system that decomposes stationary-camera video into a tile-based \emph{polyomino} data model,
enabling fine-grained spatiotemporal pruning that reduces detector calls with minimal fidelity loss.
Our evaluation across seven video datasets shows that \sys{} consistently stays within a 5\% HOTA loss on all datasets,
while prior systems exceed this bound on \comparePriorFailDatasetsFivePct{} of them.
\sys{} achieves up to \compareAccuracyMatchedSpeedupMaxFivePct{}$\times$ higher throughput than prior systems
and up to \compareNaiveSpeedupMaxFivePct{}$\times$ higher than the reference pipeline.

\label{page:maintext-end}

\clearpage

\bibliographystyle{ACM-Reference-Format}
\bibliography{references}
\label{page:references-end}

\clearpage

\ifdefined\manuscriptonly\else
\begingroup
\clearpage
\appendix
\renewcommand{\theHsubsection}{response.\arabic{subsection}}
\section*{Response to Reviewer Comments}

We deeply appreciate the time and valuable feedback the meta-reviewer and reviewers dedicated to reviewing our paper.
Your comments helped us strengthen the evaluation and explain the contributions and scope more clearly.
For the meta-reviewer's two priorities,
we added human-verified B3D labels (R1-O1) and preparation-cost and amortization results (\autoref{sec:preparation-cost}, R9-O2).
We respond to each point in the accepted revision plan below,
describing the changes and their supporting evidence.
The remaining artifact-release work is identified in our response to R7.
Changes in the manuscript are colored by reviewer:
\revRone{blue for Reviewer~1},
\revRseven{green for Reviewer~7},
and \revRnine{purple for Reviewer~9};
\revAuthor{brown} marks changes we made on our own.

\newcommand{\setm}{\setcounter{subsubsection}{0}\renewcommand{\thesubsubsection}{\textbf{R\arabic{subsection}-M\arabic{subsubsection}}}\renewcommand{\theHsubsubsection}{response.\arabic{subsection}.M.\arabic{subsubsection}}}
\newcommand{\seto}{\setcounter{subsubsection}{0}\renewcommand{\thesubsubsection}{\textbf{R\arabic{subsection}-O\arabic{subsubsection}}}\renewcommand{\theHsubsubsection}{response.\arabic{subsection}.O.\arabic{subsubsection}}}
\newcommand{\setq}{\setcounter{subsubsection}{0}\renewcommand{\thesubsubsection}{\textbf{R\arabic{subsection}-Q\arabic{subsubsection}}}\renewcommand{\theHsubsubsection}{response.\arabic{subsection}.Q.\arabic{subsubsection}}}

\setcounter{subsection}{0}
\renewcommand{\thesubsubsection}{\textbf{M\arabic{subsubsection}}}

\renewcommand{\thesubsubsection}{\textbf{R\arabic{subsection}-O\arabic{subsubsection}}}
\renewcommand{\thesubsection}{Reviewer \#\arabic{subsection}}

\subsection{\revRone{(Blue Text)}}
\seto
\subsubsection{Human annotations for B3D}
Thank you for pointing out that the B3D ground truth was not independent of the evaluated systems.
The original B3D annotations were generated using the same detector as the systems being compared.
We agree that agreement with those automatic annotations alone cannot establish tracking accuracy against independently verified trajectories.
To address this concern,
we built human-verified ground truth for the \revisionBThreeDClipCount{} B3D test clips (\revisionBThreeDFrameCount{} frames).
We generated the initial detection labels with Faster R-CNN,
which is different from RetinaNet,
the detector used by \sys{},
OTIF,
LEAP,
and the reference pipeline on B3D.
We linked these detections into tracks with OC-SORT,
which is different from the trackers used by the compared systems and the reference pipeline during track extraction.
The training code,
weights,
and the pre-labeling code are available at \url{https://anonymous.4open.science/r/tetris-b3d-fasterrcnn}.

We used a low detection-confidence threshold of 0.2 to favor recall in the initial labels.
Human reviewers then reviewed every track,
splitting tracks that cover more than one vehicle,
removing tracks that are not vehicles,
and merging fragments of the same vehicle.
Reviewers then corrected the boxes of every track whose size changes abruptly between consecutive frames.
The annotation guide,
tools,
edit sheets,
and final labels are available at \url{https://anonymous.4open.science/r/tetris-b3d-human-annotation}.
To keep the evaluation focused on the results,
we summarize the B3D annotation protocol in \autoref{sec:evaluation} and describe the review and correction steps in more detail here.
We would be happy to expand the description in the paper if the reviewers find these details helpful for interpreting the results.
We have rescored the B3D HOTA results and the reported operating-point comparisons against these human-verified labels (\autoref{sec:e2e-results}),
as well as detection accuracy after packing (\autoref{sec:packed-detection}) and the downstream query results (\autoref{sec:leap-queries}).

\subsubsection{LEAP Comparison}
Thank you for pointing out that our comparison with LEAP was not symmetric.
The original comparison did not evaluate LEAP on the queries it was designed to support and did not explain the trajectory reconstruction clearly.
We agree that evaluating LEAP on its intended queries is necessary for a fair comparison,
and that
a HOTA comparison must explain how per-frame trajectories are obtained from LEAP's output without using ground-truth or reference tracks.
We added the query evaluation in \autoref{sec:leap-queries} and clarified the trajectory reconstruction in \autoref{sec:e2e-results}.

We retain the HOTA comparison to illustrate how aggressive whole-frame sampling can reduce tracking fidelity at comparable throughput (\autoref{sec:e2e-results}).
LEAP's lower HOTA is consistent with Observation~2 in \autoref{sec:motivation}.
When entire frames are skipped,
the same sampling gap can preserve tracks in some regions while breaking associations in others.
This is why \sys{} learns a maximum sampling gap for each tile location and uses those gaps to guide polyomino pruning.
The additional query evaluation assesses LEAP on its intended workload.
A lower HOTA score therefore does not by itself imply lower accuracy on the queries that LEAP targets.

{\bf\em LEAP Track Reconstruction.}
The reviewer asked how motion patterns are assigned,
how overlapping objects are handled,
and whether reconstruction uses reference tracks.
The original wording in \autoref{sec:evaluation} about assigning the ``closest reference track'' incorrectly attributed that assignment to our evaluation procedure.
LEAP itself assigns each object a learned motion pattern and predicts its presence interval.
Our reconstructions use LEAP's detected boxes and its assigned motion pattern and predicted interval as described below.
The motion patterns are learned from training clips during LEAP's own preparation phase.

For HOTA evaluation in this revision,
we reconstruct per-frame trajectories in three ways.
(i)
We keep LEAP's detected bounding boxes on sampled frames and linearly interpolate between consecutive boxes of the same track (as we do for \sys{} and OTIF).
(ii)
We use LEAP's assigned motion pattern and linearly resample it to produce one bounding box per frame over the predicted presence interval.
(iii)
We keep the same detected boxes as in method (i) and align the assigned motion pattern to them,
matching each box's center to the closest point on the pattern.
We interpolate along the aligned pattern between consecutive detections.
In all three reconstructions,
each track keeps its LEAP-assigned identifier.
None of the three reconstructions uses ground-truth or reference-pipeline trajectories.
The rounded scores for each method are shown below:
\begin{center}
\small
\setlength{\tabcolsep}{2pt}
\begin{tabular}{@{}lrrrrrrr@{}}
Method & Amsterdam & CalDoT1 & CalDoT2 & B3D1 & B3D2 & B3D3 & B3D4 \\
\midrule
(i) & 52.6 & 51.0 & 13.4 & \textbf{16.4} & \textbf{10.5} & \textbf{9.4} & \textbf{14.6} \\
(ii) & 11.1 & 14.8 & 13.3 & 4.1 & 2.5 & 1.5 & 4.2 \\
(iii) & \textbf{52.8} & \textbf{52.4} & \textbf{22.0} & 15.7 & 10.5 & 9.4 & 14.4 \\
\end{tabular}
\end{center}
For each dataset,
the main comparison uses whichever of the three reconstruction methods gives LEAP the highest HOTA ({\bf bold}).

{\bf\em LEAP's Intended Queries.}
As the reviewer suggested,
we additionally evaluate the frame-selection,
counting,
and top-10 queries defined in the LEAP paper as a separate subsection (\autoref{sec:leap-queries}).
Frame selection returns the frames that contain at least one vehicle,
counting returns the number of vehicles in each frame,
and top-10 returns the ten frames with the most vehicles.
\autoref{sec:leap-queries} describes the setup and the operating-point selection.
We score all systems with the query functions from LEAP's original evaluator.

The results in \autoref{sec:leap-queries} support the reviewer's point that HOTA alone understates LEAP on its intended workload.
Among the three track-materializing systems,
on frame selection,
LEAP is within 1.4 points of the best system on Amsterdam and CalDoT1.
On B3D,
every ground-truth frame contains a vehicle,
so all three systems reach at least 99\% F1 and frame selection does not distinguish them.
LEAP is also best or tied on top-10 for two of seven datasets.
Its counting error is the highest on six of seven datasets.
Counting penalizes every per-frame count error,
so each missed,
extra,
or mistimed object adds error on every frame it affects.
Frame selection penalizes only errors that change whether a frame contains a vehicle.
A missed vehicle therefore costs nothing on a frame where another vehicle is still tracked,
and an extra vehicle costs nothing on a frame that already contains one.
A system can therefore be close to the best on frame selection while having the highest counting error,
as LEAP is on Amsterdam and CalDoT1.

Overall,
despite the new reconstructions and the added queries,
\sys{} remains more accurate than LEAP.
On every dataset,
a \sys{} configuration with higher HOTA than LEAP
runs \revisionSpeedupLeapmatchedLeapMin{}--\revisionSpeedupLeapmatchedLeapMax{}$\times$ faster than LEAP.
On LEAP's
queries,
\sys{} runs faster and is at least as accurate as LEAP in all \revisionQueryLeapCases{}
cases,
and more accurate in \revisionQueryLeapBetterCount{} of them.

\subsubsection{Runtime accounting and scalability}
Thank you for raising the question of end-to-end cost and ILP scalability.
The original submission did not make clear whether the reported runtime includes classification and ILP solving,
did not report the cost of the classifier or the ILP on their own,
and did not report the cost of the training stage.
We agree that these costs are necessary to assess whether \sys{} remains practical for longer or higher-resolution video.
To address this concern,
we clarified which operators the reported extraction time includes
and added 3 new measurements.

Classification and ILP solving were already included in the reported extraction time,
but our previous explanation did not make that clear.
The experiment setup (\autoref{sec:evaluation}) now lists every operator in the extraction time,
which
excludes preparation and disk I/O.

First,
we added a per-operator runtime breakdown (\autoref{sec:ilp-cost}, \autoref{fig:runtime-breakdown}).
It runs one fixed configuration on every dataset and times each extraction operator separately,
so the costs of classification and ILP pruning are reported on their own.
We also ran the classifier and the reference detector on the same frames on the same GPU to compare their costs directly (please see also our response to R9-O1).

Second,
we added an ILP scalability study (\autoref{sec:ilp-scalability}).
It reports the size of each ILP and how many finish within \sys{}'s \revisionFixedIlpBudgetSeconds{}\,s solver budget,
which \autoref{sec:prune} now states.
To measure the pruning that this budget leaves unrealized,
we reran all \revisionFixedIlpRuns{} ILPs with a 1-hour budget (please see also our response to R9-O3).
We show that the longer budget removes only \revisionFixedIlpHourAllTilesReducedPct{}\% more of the tiles kept under the \revisionFixedIlpBudgetSeconds{}\,s budget.
With this budget,
the whole pruning step,
including ILP construction and solving,
takes only \revisionFixedPruningPipelinePctMin{}--\revisionFixedPruningPipelinePctMax{}\% of extraction time.

Third,
we measured the preparation cost of \sys{} and both baselines,
and how much video is needed to amortize this cost (\autoref{sec:preparation-cost}).
When preparation is charged before extraction,
\sys{} recovers its cost within \revisionCrossTetrisLatestHours{} elapsed hours on every dataset,
after which a longer video only adds to its advantage over the reference pipeline and OTIF
(please see also our response to R9-O2).

{\bf\em Resolution scaling.}
All datasets are at most 1280$\times$720 pixels.
The classifier runs on each tile,
so its cost grows with the number of tiles,
while the detector runs on each frame or packed canvas.
To compare how the two costs grow,
we ran both on frames with 2$\times$ and 4$\times$ the pixels of each dataset's frames,
with
the largest batch that fits in GPU memory.
Both times grow at similar rates as the number of pixels increases,
so they stay proportional to each other:
\begin{center}
\small
\begin{tabular}{@{}lrrrrrr@{}}
& \multicolumn{2}{c}{Detection} & \multicolumn{2}{c}{Classification} & \multicolumn{2}{c}{Classification / detection} \\
\cmidrule(lr){2-3}\cmidrule(lr){4-5}\cmidrule(l){6-7}
Pixels & 2$\times$ & 4$\times$ & 2$\times$ & 4$\times$ & 2$\times$ & 4$\times$ \\
\midrule
Ams. & 1.99$\times$ & 3.85$\times$ & 1.90$\times$ & 3.79$\times$ & 0.96 & 0.98 \\
CDoT1 & 1.80$\times$ & 3.48$\times$ & 1.98$\times$ & 3.88$\times$ & 1.10 & 1.12 \\
CDoT2 & 1.77$\times$ & 3.42$\times$ & 1.99$\times$ & 3.90$\times$ & 1.12 & 1.14 \\
B3D & 1.87$\times$ & 4.07$\times$ & 1.79$\times$ & 3.77$\times$ & 0.96 & 0.93 \\
\end{tabular}
\end{center}
At every tested resolution,
the detector takes \revisionResolutionRatioMin{}--\revisionResolutionRatioMax{}$\times$ as long as the classifier,
so the classifier's overhead stays proportional to the detector's cost.
We did not include this experiment in the paper due to space constraints.
If the reviewer finds these results important for evaluating our system,
we are happy to add them.

\subsubsection{Detection accuracy after packing}
Thank you for raising the concern about how packing changes the detector's input distribution.
Packing efficacy alone
did not establish how packing affects detection accuracy.
Packing polyominoes
into canvases introduces
context that differs from the natural full frames the detector was trained on.
We agree that a direct comparison of detection accuracy before and after packing is necessary.
To address this concern,
we add precision and recall of the reference detector on original frames and packed canvases (\autoref{sec:packed-detection},
\autoref{fig:classifier-and-packing}).

On every dataset,
packing changes precision and recall by at most \revisionPackingDeployedMaxChange{} percentage points on the specified configurations in \autoref{sec:packed-detection}.
Without padding ($P$=0),
packing lowers recall by \revisionPackingSparseNoPadRecallDropMin{}--\revisionPackingSparseNoPadRecallDropMax{} points on Amsterdam and CalDoT,
but changes precision and recall by at most \revisionPackingDenseMaxChange{} points on B3D.
Padding trades detection throughput for this accuracy.
On Amsterdam and CalDoT,
padding
recovers recall to within \revisionPackingSparsePadRecallMaxChange{} points of the original detection,
but raises detector calls from \revisionPaddingSparseNoPadCallsMin{}--\revisionPaddingSparseNoPadCallsMax{}\% to \revisionPaddingSparsePadCallsMin{}--\revisionPaddingSparsePadCallsMax{}\% of the original pipeline's calls.
On B3D,
padding barely changes accuracy but raises detector calls from \revisionPaddingDenseNoPadCallsMin{}--\revisionPaddingDenseNoPadCallsMax{}\% to \revisionPaddingDensePadCallsMin{}--\revisionPaddingDensePadCallsMax{}\%,
approaching detection on every frame:
\begin{center}
\small
\setlength{\tabcolsep}{3pt}
\resizebox{.99\columnwidth}{!}{%
\begin{tabular}{@{}lrrrrrrr@{}}
& Ams. & CDoT1 & CDoT2 & B3D1 & B3D2 & B3D3 & B3D4 \\
\midrule
Recall change ($P$=0) & $-$7.7 & $-$3.2 & $-$9.9 & $-$0.7 & 0.0 & 0.6 & 1.4 \\
Recall change ($P$=1) & $-$0.5 & 0.0 & 0.1 & 0.2 & $-$0.3 & $-$0.3 & 0.2 \\
Detector calls, \% of original ($P$=0) & 4.2 & 7.5 & 4.5 & 18.8 & 29.8 & 16.6 & 12.4 \\
Detector calls, \% of original ($P$=1) & 12.3 & 25.0 & 20.0 & 71.8 & 91.2 & 65.8 & 45.2 \\
\end{tabular}%
}
\end{center}
The trade-off padding provides between accuracy and throughput
therefore depends on the dataset and its object patterns,
and \sys{}'s Pareto-frontier search selects the appropriate padding for each dataset.
Since preserving object scale does not always guarantee unchanged detector accuracy on all scenarios,
we removed the claim that packed detection stays accurate because canvases preserve object scale from \autoref{sec:exec}.
We did not include this padding analysis in the paper due to space constraints.
If the reviewer finds it important for evaluating our system,
we are happy to add it.

\subsubsection{Figure readability}
Thank you for pointing out the hard-to-read figures and for suggesting a tolerance-to-gap curve.
In the original submission,
Figure~7 showed the tolerance-to-gap mapping only as matrices.
We agree that a tolerance-to-gap curve makes it easier to see how each tolerance determines the sampling gaps.
\autoref{fig:missrate-to-g} now includes a tolerance-to-gap curve, showing
how each tolerance maps to the largest sampling gap for each tile position.
Also, \autoref{fig:track-by-detect} no longer has text occlusion.

\setm
\subsubsection{Tile Size}
Thank you for pointing out the missing tile size explanation.
We agree that the tile size and its rationale,
missing from the original submission,
should be explicit.
\autoref{sec:dm-tiles} now states the tile size ($TS$=60) and its trade-off.
In early experiments,
$TS$=60 gave a better trade-off than $TS$=30 or $TS$=120 on every dataset.

\subsubsection{Camera and distribution changes}
Thank you for raising the question of generalizing across cameras and conditions.
The original submission trained,
validated,
and tested within each stationary-camera dataset
and did not state how \sys{} handles other cameras or conditions.
We agree that the paper should state this scope.
\autoref{sec:introduction} (in purple) now defines a dataset as videos from a single camera
and states that changing the dataset,
detector,
or tracker requires repeating the affected preparation.
\autoref{sec:overview} adds that cross-camera transfer and robustness to weather or traffic shifts are not evaluated.
OTIF and LEAP likewise prepare for each dataset.

\setcounter{subsection}{6}
\subsection{\revRseven{(Green Text)}}
\seto
Thank you for the detailed comments and for the pointers to related work in other communities.
The reviewer's comments share one major concern,
that \sys{}'s techniques resemble established work in video processing.
We believe that \sys{} makes a distinct contribution,
but our original manuscript did not convey it.
Rereading it with these comments,
we found two weaknesses.
The writing did not lay out the landscape of video analytics systems well enough to distinguish \sys{} from existing work,
and the evaluation stopped at track materialization and did not extend to the query processing built on it.
The revision addresses the writing in the related work (\autoref{sec:related-systems}) and the data model (\autoref{sec:data-model}),
and the evaluation with a query-processing comparison (\autoref{sec:leap-queries}).

\subsubsection{Contribution and closest prior systems}
Thank you for asking us to distinguish \sys{} from prior systems in the multimedia and networking literature.
The original submission did not explain how \sys{} relates to region selection,
motion-guided sampling,
and packing in the multimedia and networking literature.
We agree that these techniques are established.
To address this concern,
we added
\autoref{sec:related-systems} to clearly explain that video analytics systems share such techniques but specialize them to different tasks,
scene assumptions,
and pipeline stages,
and that studying them across these specializations helps identify their strengths.
\autoref{sec:related-systems} also discusses RegenHance,
MOSAIC,
How2Compress,
and Crucio.
\sys{}'s specialization is
materializing object tracks from stationary video,
where pruning must preserve the trajectories that downstream queries reuse.
While we share region grouping and packing with RegenHance and MOSAIC,
our contribution is in using polyominoes as units of joint spatial and temporal selection,
guided by a maximum sampling gap learned for each tile location.
The ``Optimization Operators'' ablation (\autoref{sec:ablation-operators}) evaluates the benefit of the combined operators.
Under a 5\% HOTA loss,
adding polyomino pruning to relevance classification and packing raises the average speedup over the reference pipeline from \ablationOneSpeedupAvg{}$\times$ to \ablationTwoSpeedupAvg{}$\times$.

\subsubsection{Polyominoes as optimization units}
Thank you for raising the apparent similarity between our data model and object segmentation.
The original submission did not make clear that polyominoes are not segmentation results.
We agree that the role of polyominoes should be clear.
To address this concern,
\autoref{sec:data-model} now introduces tiles and polyominoes as units of optimization for faster tracking,
and \autoref{sec:dm-polyominoes} now states that a polyomino may contain multiple objects and does not represent an object mask or identity.
\sys{}'s optimizer prunes and packs each polyomino as a whole,
so retaining one polyomino can satisfy the sampling requirements of all the tile locations it contains.
Unlike sampling driven by motion vectors,
\sys{} learns each tile location's maximum sampling gap from the supplied tracker's mistrack rates (\autoref{sec:learn-constraints}).
Although the observations in \autoref{sec:motivation} are not new findings on their own,
we use them to motivate \sys{}'s tile-level relevance classification,
per-tile pruning,
and polyomino grouping,
which together speed up track materialization.
We have added Glimpse (SenSys) and PreActo (PerCom) to \autoref{sec:related-systems},
explaining their offloading and cross-camera handover settings and how these optimizations can complement \sys{}'s reduction of detector input.

\subsubsection{Materialization and supported queries}
Thank you for the questions on query processing over the materialized tracks,
semantic queries,
and alternative optimization approaches.
The reviewer found our query processing naive,
since the original submission did not make clear that \sys{}'s contribution lies in track extraction rather than in new query operators.
We agree that the database angle needs to be developed in more detail,
including the queries that run over the materialized tracks.
\sys{} makes it cheap to materialize detailed object tracks once,
so that many track queries can then run over the stored tracks (\autoref{sec:introduction},
\autoref{sec:track-materialization}).
To evaluate how accurately the materialized tracks answer queries,
we compare \sys{} against OTIF and LEAP on frame-selection,
counting,
and top-10 queries defined by LEAP (\autoref{sec:leap-queries}).
\sys{} performs best
in most cases.

For the reviewer's question on semantic queries,
the materialized tracks support approximate queries over positions,
identities,
classes,
and time.
Semantic features such as appearance embeddings could be materialized alongside the tracks by adding an extraction operator,
which we leave to future work.
We also stated \sys{}'s scope more clearly in the ``specialization and scope'' of \autoref{sec:overview} (in blue).

The reviewer also suggested the group-of-frames approach,
which identifies scene changes,
and asked us to update the baselines with work from other venues.
We agree that the evaluation should include such systems.
To address this concern,
we added NoScope and Seiden as baselines in the query evaluation of \autoref{sec:leap-queries}
and discuss both in \autoref{sec:related-systems}.
Seiden samples frames with a multi-armed bandit that prioritizes segments where query results vary,
and NoScope detects changes between frames to skip redundant ones.
Both sample whole frames and output query answers directly,
whereas \sys{} samples tiles and preserves per-object tracks.
OTIF and LEAP remain our closest tracking baselines because,
like \sys{},
they extract every object's track in a single-device setting.
Only these two systems remain in our end-to-end evaluation (\autoref{sec:e2e-results}),
which measures tracking accuracy (HOTA),
since NoScope and Seiden do not output tracks.

NoScope supports only frame selection,
so we evaluate it on that query alone.
To make the accuracy comparison fair,
we run NoScope and Seiden at operating points with throughput comparable to LEAP's and lower than \sys{}'s.
To find these operating points,
we sweep each system's tradeoff between throughput and accuracy.
For NoScope,
we tune its cascade at each target error rate in its default sweep.
For Seiden,
we sweep its frame budget over 1/4, 1/8, 1/12, \ldots, 1/32 of the test frames.
Against all 4 baselines,
\sys{} is the best in \revisionQueryAllBestPct{}\% of the \revisionQueryLeapCases{} cases
and within \revisionQueryAllWithinFiveMaxGapPp{} points in \revisionQueryAllWithinFivePct{}\%,
trailing Seiden by \revisionQueryAllLowerSecondGapPp{} and \revisionQueryAllLowerMaxGapPp{} points in the other 2,
on B3D counting.

\renewcommand{\thesubsubsection}{\textbf{R\arabic{subsection}-12}}
\renewcommand{\theHsubsubsection}{response.\arabic{subsection}.12}
\subsubsection{Artifacts and reproduction}
Thank you for asking us to clarify the implementation and evaluation setting.
The original submission did not state the evaluation setting clearly or explain how to reproduce the results.
We agree that both should be explicit.
The specialization and scope paragraph of \autoref{sec:overview} (in blue) and the repository now state that \sys{} processes stored video on a single server without distributed camera coordination.
The artifact implements the complete preparation and extraction pipeline.

We have improved the repository's reproduction guide with the dependency and solver requirements,
the data preparation steps,
and the definitions of the training,
validation,
and test splits.
The repository also provides a complete example from preparation through track extraction,
and commands that link each reported figure and table to the scripts that generate it.
We also release the instructions and tooling we used to build the human-verified B3D labels (please see our response to R1-O1).

\setcounter{subsection}{8}
\subsection{\revRnine{(Purple Text)}}
\seto
\subsubsection{Relevance classifier cost}
Thank you for raising whether the relevance classifier is cheap enough to run on every frame.
The original submission did not compare its cost with the detector's.
We agree that this comparison is needed.
To address this concern,
\autoref{sec:motivation} now notes that a small classifier over tiles identifies irrelevant tiles at a fraction of the detector's cost.
The classifier is a small ShuffleNet (\autoref{sec:relevance-classification}),
and its cost was already included in the reported extraction time (please see our response to R1-O3).
\autoref{sec:ilp-cost} now reports that classifying every frame costs only \revisionFixedClassifierDetectorPctMin{}--\revisionFixedClassifierDetectorPctMax{}\% of running the detector on every frame.

For this response,
we also measured throughput on the same frames and GPU for each of the seven datasets.
The classifier processes the tiles of each frame at \revisionClassifierFpsMin{}--\revisionClassifierFpsMax{} FPS,
while the detector processes frames at \revisionDetectorFpsMin{}--\revisionDetectorFpsMax{} FPS.

\subsubsection{Preparation cost and amortization}
Thank you for raising the question regarding \sys{}'s preparation cost.
The original submission did not report these costs,
including the detector runs used to train the classifier.
We agree that these costs are needed to assess when the extraction speed benefit outweighs the initial preparation overhead.
To address this concern,
\autoref{sec:preparation-cost} now reports the preparation time of \sys{},
OTIF,
and LEAP on every dataset,
and when \sys{} overtakes each of them after both preparation and extraction.
During training,
the detector runs only once over the training split.
Its output supplies both the classifier's labels (\autoref{sec:specialization}) and the reference tracks for mistrack profiling (\autoref{sec:learn-constraints}),
and profiling each candidate gap reruns only the inexpensive tracker.
The table below breaks down of \sys{}'s preparation time in hours.
The validation sweep dominates,
while the detector pass,
classifier training,
and mistrack profiling take far less time:
\begin{center}
\small
\setlength{\tabcolsep}{3pt}
\resizebox{.99\columnwidth}{!}{%
\begin{tabular}{@{}lrrrrrrr@{}}
& Ams. & CDoT1 & CDoT2 & B3D1 & B3D2 & B3D3 & B3D4 \\
\midrule
Detector pass (training split) & 0.9 & 0.4 & 0.4 & 0.1 & 0.1 & 0.1 & 0.1 \\
Classifier training & 6.3 & 2.5 & 2.2 & 0.9 & 0.7 & 0.9 & 0.8 \\
Mistrack profiling & 0.1 & 0.1 & 0.1 & $<$0.1 & $<$0.1 & $<$0.1 & $<$0.1 \\
Validation sweep & 24.7 & 14.3 & 17.0 & 7.5 & 8.5 & 7.2 & 5.2 \\
\midrule
Total (hours) & 31.9 & 17.3 & 19.7 & 8.6 & 9.3 & 8.3 & 6.1 \\
\end{tabular}%
}
\end{center}
The progress curves in \autoref{fig:amortization} combine these totals with extraction throughput
to project when \sys{} overtakes each baseline,
including the reference pipeline.

\subsubsection{ILP scalability}
Thank you for raising the question of ILP scalability.
The original submission did not report ILP sizes,
solve times,
or the stopping criterion.
We agree that these are needed to assess how the ILP scales.
To address this concern,
\autoref{sec:prune} now states the \revisionFixedIlpBudgetSeconds{}-s solver budget and that \sys{} uses its best feasible solution.

\autoref{sec:ilp-cost} times each extraction operator separately for one fixed configuration on every dataset.
Before presolve,
the largest ILP has \revisionFixedIlpVariablesMax{} variables and \revisionFixedIlpConstraintsMax{} constraints,
and pruning,
including ILP construction and solving,
takes only \revisionFixedPruningPipelinePctMin{}--\revisionFixedPruningPipelinePctMax{}\% of extraction time,
which the reported runtime for our end-to-end evaluation already includes.
We set the solver budget to \revisionFixedIlpBudgetSeconds{} seconds per 1-minute video clip and accept its best feasible solution.
At this budget,
the measurements in \autoref{sec:ilp-cost} show that pruning takes much less time than classification.
Since the
ILP solving
cost is bounded per one-minute clip,
the ILP will not bottleneck \sys{}'s track extraction.

To show that our \revisionFixedIlpBudgetSeconds{}-s budget is sufficient,
\autoref{sec:ilp-scalability} reruns all \revisionFixedIlpRuns{} ILPs with a 1-hour budget to measure the pruning that the short budget gives up.
The 1-hour budget prunes only \revisionFixedIlpHourAllTilesReducedPct{}\% more tiles,
and Gurobi's lower bound shows that even an optimal solution could prune at most \revisionFixedIlpHourBoundPct{}\% more,
so the short budget loses little pruning.

\setq
\subsubsection{Detector compatibility and retraining}
Thank you for challenging our detector-agnostic claim.
The original submission used ``detector-agnostic'' without saying what changing the detector requires.
We agree that accepting a detector interface does not imply that the learned classifier transfers to a different detector.

To answer the question,
yes,
changing the detector requires retraining the classifier,
because its labels come from that detector's outputs (\autoref{sec:specialization}).
The sampling gaps depend on both the detector and the tracker,
so they are profiled again.
The configurations are then validated again to regenerate the Pareto frontier.

By ``detector-agnostic,''
we mean that the data model and execution engine work with any compatible detector without modification,
as our evaluation shows with YOLOv5 and RetinaNet.
Like OTIF and LEAP,
\sys{} prepares for each target dataset,
and this preparation is a one-time cost for each detector rather than a cost for each video.
The cost is amortized as more video is processed.
\sys{} recovers its full preparation time within \revisionCrossTetrisLatestHours{} elapsed hours on every dataset,
after which it processes video \revisionSpeedupTetrisReferenceMin{}--\revisionSpeedupTetrisReferenceMax{}$\times$ faster than the reference pipeline.
\autoref{sec:introduction} and \autoref{sec:overview} now state explicitly that changing the detector or tracker requires repeating this preparation,
with the \autoref{sec:overview} paragraph in R1's color (blue) since it also answers R1-M2.

\setm
\renewcommand{\thesubsubsection}{\textbf{R\arabic{subsection}-M1/M3}}
\renewcommand{\theHsubsubsection}{response.\arabic{subsection}.M.13}
\subsubsection{Figure readability}
Thank you for pointing out the issues in Figure~1 and Figure~9.
\autoref{fig:track-by-detect} now shows no text occlusion,
and \autoref{fig:classifier-and-packing} (previously Figure~9) now has larger labels and axes.

\setm
\setcounter{subsubsection}{1}
\subsubsection{Forward references in Section 5}
Thank you for pointing out that our overused forward references hurt the readability of \autoref{sec:exec}.
We agree that these forward references made \autoref{sec:exec} harder to follow.
We reduced the forward references in \autoref{sec:exec} from 8 to 2.
Since \autoref{sec:overview} already introduces the relevance classifier and the maximum sampling gaps,
\autoref{sec:exec} now treats them as inputs instead of pointing ahead to their full explanation in \autoref{sec:e2e}.
We also moved the table of system configuration parameters (\autoref{tab:frontier-sweep}) from \autoref{sec:e2e} to \autoref{sec:overview},
where we describe Pareto frontier generation
and introduce every configuration parameter that readers need to understand \autoref{sec:exec}.

\subsection*{Removal}
To make space for the new results,
\autoref{fig:e2e-results} no longer includes the rows for \sys{}'s speedup at matched HOTA and HOTA gain at matched throughput;
however, \autoref{sec:e2e-results} still reports both comparisons.
The table of detection's share of the reference pipeline's compute is replaced by the range stated in \autoref{sec:tracking-by-detection}.
In \autoref{fig:classifier-and-packing},
we removed the classifier's precision and recall panels,
which we do not discuss;
the classifier's F1 combines both metrics.

\endgroup
\fi

\clearpage
\appendix
\section{NP-hardness Proofs}
\label{sec:np-proofs}
We now prove the NP-hardness of the two optimization problems described in \autoref{sec:exec}:
polyomino pruning (\autoref{sec:prune}) and polyomino packing (\autoref{sec:pack}).

\subsection{Polyomino Pruning is NP-hard}
\label{sec:np-pruning}

Recall the pruning problem from \autoref{sec:prune}.
Given $N$ frames of relevant polyominoes and a maximum tile sampling gap $\bar{g}^{\bar{M}}_{i,j}$ for each tile location,
we select a subset of the polyominoes that minimizes the total number of selected tiles,
such that every window of $\bar{g}^{\bar{M}}_{i,j}$ consecutive frames containing a polyomino that covers $(i, j)$ also contains a selected polyomino covering $(i, j)$.

We prove NP-hardness in two stages.
First (\autoref{sec:np-relaxed-encodes-vc}),
we define a relaxed variant of polyomino pruning
in which some tiles do not enforce a sampling gap,
denoted $\bar{g}^{\bar{M}}_{i,j} = \textnormal{null}$;
such tiles need not be covered by any selected polyomino.
We show that this relaxed variant can encode the Minimum Vertex
Cover problem on planar graphs with maximum degree of three,
which is NP-hard~\cite{garey1977rectilinear,garey1979computers}.
Second (\autoref{sec:np-encodes-relaxed}),
we show that polyomino pruning as originally formulated,
where every tile enforces a sampling gap,
can encode the relaxed variant,
so the hardness carries over.

\subsubsection{Relaxed Polyomino Pruning Encodes Minimum Vertex Cover}
\label{sec:np-relaxed-encodes-vc}
In the pruning instance that the reduction constructs,
selecting a polyomino corresponds to placing a vertex in the cover,
one tile per edge carries a sampling gap that only the polyominoes of the edge's two endpoints can satisfy,
and the minimum tile objective minimizes the number of selected vertices.

\begin{figure}[t]
    \centering
    \resizebox{\columnwidth}{!}{\begin{tikzpicture}[x=0.5cm,y=0.5cm]
  \tikzstyle{labelstyle}=[font=\footnotesize]
  \tikzstyle{tinylabel}=[font=\tiny]
  \tikzstyle{guidestyle}=[line width=0.3pt, draw=black!30]
  \tikzstyle{pathtile}=[draw=black!30, fill=black!8, line width=0.3pt]
  \tikzstyle{verttile}=[draw=black!70, fill=black!60, line width=0.3pt]
  \tikzstyle{midtile}=[draw=orange!85!black, fill=orange!30, line width=0.45pt]

  \begin{scope}[shift={(0.5,2.3)}]
    \node[labelstyle, anchor=west] at (-0.6,4.6) {Input graph $G$};
    \draw[line width=0.6pt] (0,3) -- (3,3);
    \draw[line width=0.6pt] (0,3) -- (0,0);
    \draw[line width=0.6pt] (3,3) -- (0,0);
    \draw[line width=0.6pt] (3,3) -- (3,0);
    \draw[line width=0.6pt] (0,0) -- (3,0);
    \draw[line width=0.6pt] (3,0) -- (6,0);
    \foreach \x/\y/\n/\ax/\ay in {0/3/a/-0.45/0, 3/3/b/0.45/0.3, 0/0/c/-0.45/0, 3/0/d/0/-0.5, 6/0/e/0.45/0} {
      \fill[black!60] (\x,\y) circle (0.24);
      \node[labelstyle] at (\x+\ax,\y+\ay) {$\n$};
    }
  \end{scope}

  \begin{scope}[shift={(7.5,0)}]
    \node[labelstyle, anchor=west] at (0.8,8.7) {Layout on the tile grid};
    \foreach \x/\y in {5/5, 3/3, 6/3, 4/2, 8/2,
                       6/6,6/7,5/7,4/7,3/7,2/7,1/6,1/5,1/4,1/3,1/2,2/2} {
      \draw[pathtile] (\x,\y) rectangle ++(1,1);
    }
    \node[tinylabel, text=black!55] at (3.5,7.5) {$bc$};
    \foreach \x/\y in {4/5, 3/4, 6/4, 5/2, 7/2, 1/7} {
      \draw[midtile] (\x,\y) rectangle ++(1,1);
    }
    \foreach \x/\y/\n in {3/5/a, 6/5/b, 3/2/c, 6/2/d, 9/2/e} {
      \draw[verttile] (\x,\y) rectangle ++(1,1);
      \node[labelstyle, text=white] at (\x+0.5,\y+0.5) {$\n$};
    }
    \draw[guidestyle, step=1] (1,2) grid (10,8);
  \end{scope}
\end{tikzpicture}}
    \Description{Left: a planar graph with five vertices a through e and six edges.
    Right: the same graph on a tile grid,
    with dark vertex tiles, light edge-path tiles including one path with bends, and one orange midpoint tile per edge.}
    \caption{Laying a graph out on the tile grid.
    \panellabel{Left} A planar input graph with maximum degree three.
    \panellabel{Right} Its layout.
    Each vertex occupies a tile,
    each edge becomes a path of tiles routed along grid rows and columns,
    possibly with bends (edge $bc$ is routed around the top-left of the grid, bending at its corners),
    and distinct edge paths share no tiles.
    One interior tile of each edge path is marked as the edge's midpoint (orange).}
    \label{fig:np-layout}
\end{figure}
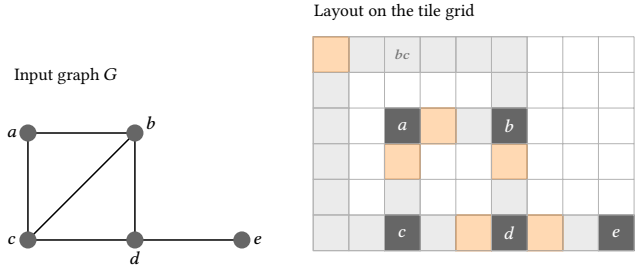

\paragraph{Step 1: The graph as a tile grid.}
Given such a graph $G = (V, E)$ with $n = |V|$ vertices,
we lay the graph out on a tile grid whose dimensions $R \times C$ are polynomial in $n$,
with edges routed along grid lines~\cite{valiant1981universality},
as shown in \autoref{fig:np-layout}.
Each vertex occupies one tile,
each edge becomes a path of tiles between its two endpoint vertices' tiles,
and every edge path occupies at least one tile strictly between its two endpoints;
we mark one such interior tile $m_e$ on each edge path as the \emph{midpoint} of edge $e$.
This layout represents the whole graph spatially on a single tile grid of size $R \times C$.

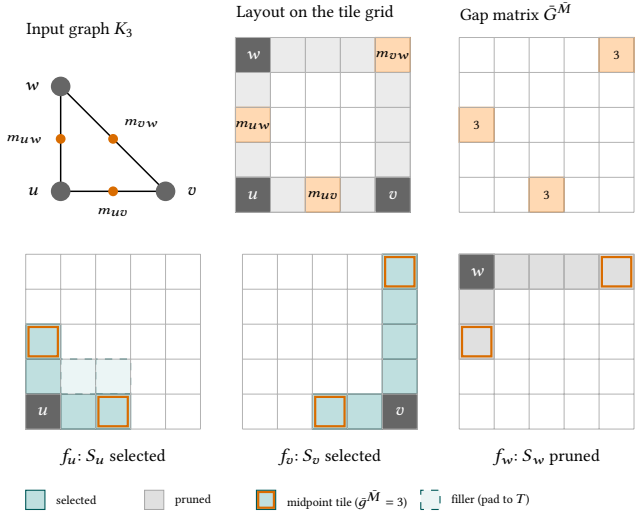
\begin{figure}[t]
    \centering
    \resizebox{\columnwidth}{!}{\begin{tikzpicture}[x=0.5cm,y=0.5cm]
  \tikzstyle{labelstyle}=[font=\footnotesize]
  \tikzstyle{tinylabel}=[font=\tiny]
  \tikzstyle{guidestyle}=[line width=0.3pt, draw=black!30]
  \tikzstyle{pathtile}=[draw=black!30, fill=black!8, line width=0.3pt]
  \tikzstyle{verttile}=[draw=black!70, fill=black!60, line width=0.3pt]
  \tikzstyle{keeptile}=[draw=teal!70!black, fill=teal!25, line width=0.45pt]
  \tikzstyle{droptile}=[draw=black!35, fill=black!12, line width=0.3pt]
  \tikzstyle{fillertile}=[draw=teal!70!black, dashed, fill=teal!8, line width=0.45pt]
  \tikzstyle{midtile}=[draw=orange!85!black, fill=orange!30, line width=0.45pt]
  \tikzstyle{midframe}=[draw=orange!85!black, fill=none, line width=0.9pt]

  \begin{scope}[shift={(1.0,6.8)}]
    \node[labelstyle, anchor=west] at (-1.2,4.6) {Input graph $K_3$};
    \draw[line width=0.6pt] (0,0) -- (3,0) -- (0,3) -- cycle;
    \fill[orange!85!black] (1.5,0) circle (0.14);
    \fill[orange!85!black] (0,1.5) circle (0.14);
    \fill[orange!85!black] (1.5,1.5) circle (0.14);
    \node[tinylabel, anchor=north] at (1.5,-0.25) {$m_{uv}$};
    \node[tinylabel, anchor=east] at (-0.3,1.5) {$m_{uw}$};
    \node[tinylabel, anchor=south west] at (1.6,1.6) {$m_{vw}$};
    \fill[black!60] (0,0) circle (0.28);
    \fill[black!60] (3,0) circle (0.28);
    \fill[black!60] (0,3) circle (0.28);
    \node[labelstyle, anchor=east] at (-0.4,0) {$u$};
    \node[labelstyle, anchor=west] at (3.4,0) {$v$};
    \node[labelstyle, anchor=east] at (-0.4,3) {$w$};
  \end{scope}

  \begin{scope}[shift={(6.0,6.2)}]
    \node[labelstyle, anchor=west] at (-0.2,5.7) {Layout on the tile grid};
    \foreach \x/\y in {1/0,3/0,0/1,0/3,4/1,4/2,4/3,3/4,2/4,1/4} {
      \draw[pathtile] (\x,\y) rectangle ++(1,1);
    }
    \foreach \x/\y in {2/0,0/2,4/4} {
      \draw[midtile] (\x,\y) rectangle ++(1,1);
    }
    \foreach \x/\y/\n in {0/0/u,4/0/v,0/4/w} {
      \draw[verttile] (\x,\y) rectangle ++(1,1);
      \node[labelstyle, text=white] at (\x+0.5,\y+0.5) {$\n$};
    }
    \node[tinylabel] at (2.5,0.5) {$m_{uv}$};
    \node[tinylabel] at (0.5,2.5) {$m_{uw}$};
    \node[tinylabel] at (4.5,4.5) {$m_{vw}$};
    \draw[guidestyle, step=1] (0,0) grid (5,5);
  \end{scope}

  \begin{scope}[shift={(12.4,6.2)}]
    \node[labelstyle, anchor=west] at (-0.2,5.7) {Gap matrix $\bar{G}^{\bar{M}}$};
    \foreach \x/\y in {2/0,0/2,4/4} {
      \draw[midtile] (\x,\y) rectangle ++(1,1);
      \node[tinylabel] at (\x+0.5,\y+0.5) {$3$};
    }
    \draw[guidestyle, step=1] (0,0) grid (5,5);
  \end{scope}

  \begin{scope}[shift={(0,0)}]
    \foreach \x/\y in {1/0,0/1} {
      \draw[keeptile] (\x,\y) rectangle ++(1,1);
    }
    \foreach \x/\y in {1/1,2/1} {
      \draw[fillertile] (\x,\y) rectangle ++(1,1);
    }
    \foreach \x/\y in {2/0,0/2} {
      \draw[keeptile] (\x,\y) rectangle ++(1,1);
      \draw[midframe] (\x+0.08,\y+0.08) rectangle ++(0.84,0.84);
    }
    \draw[verttile] (0,0) rectangle ++(1,1);
    \node[labelstyle, text=white] at (0.5,0.5) {$u$};
    \draw[guidestyle, step=1] (0,0) grid (5,5);
    \node[labelstyle] at (2.5,-0.8) {$f_u$: $S_u$ selected};
  \end{scope}

  \begin{scope}[shift={(6.2,0)}]
    \foreach \x/\y in {3/0,4/1,4/2,4/3} {
      \draw[keeptile] (\x,\y) rectangle ++(1,1);
    }
    \foreach \x/\y in {2/0,4/4} {
      \draw[keeptile] (\x,\y) rectangle ++(1,1);
      \draw[midframe] (\x+0.08,\y+0.08) rectangle ++(0.84,0.84);
    }
    \draw[verttile] (4,0) rectangle ++(1,1);
    \node[labelstyle, text=white] at (4.5,0.5) {$v$};
    \draw[guidestyle, step=1] (0,0) grid (5,5);
    \node[labelstyle] at (2.5,-0.8) {$f_v$: $S_v$ selected};
  \end{scope}

  \begin{scope}[shift={(12.4,0)}]
    \foreach \x/\y in {0/3,1/4,2/4,3/4} {
      \draw[droptile] (\x,\y) rectangle ++(1,1);
    }
    \foreach \x/\y in {0/2,4/4} {
      \draw[droptile] (\x,\y) rectangle ++(1,1);
      \draw[midframe] (\x+0.08,\y+0.08) rectangle ++(0.84,0.84);
    }
    \draw[verttile] (0,4) rectangle ++(1,1);
    \node[labelstyle, text=white] at (0.5,4.5) {$w$};
    \draw[guidestyle, step=1] (0,0) grid (5,5);
    \node[labelstyle] at (2.5,-0.8) {$f_w$: $S_w$ pruned};
  \end{scope}

  \begin{scope}[shift={(0,-2.3)}]
    \draw[keeptile] (0,0) rectangle ++(0.55,0.55);
    \node[tinylabel, anchor=west] at (0.65,0.27) {selected};
    \draw[droptile] (3.4,0) rectangle ++(0.55,0.55);
    \node[tinylabel, anchor=west] at (4.05,0.27) {pruned};
    \draw[keeptile] (6.6,0) rectangle ++(0.55,0.55);
    \draw[midframe] (6.65,0.05) rectangle ++(0.45,0.45);
    \node[tinylabel, anchor=west] at (7.25,0.27) {midpoint tile ($\bar{g}^{\bar{M}} = 3$)};
    \draw[fillertile] (11.3,0) rectangle ++(0.55,0.55);
    \node[tinylabel, anchor=west] at (11.95,0.27) {filler (pad to $T$)};
  \end{scope}
\end{tikzpicture}}
    \Description{Top left: the triangle graph on vertices u, v, and w with a midpoint marked on each edge.
    Top middle: the same graph drawn on a five-by-five tile grid,
    with dark vertex tiles, light edge-path tiles, and three orange midpoint tiles.
    Top right: the gap matrix as a five-by-five grid with the value 3 on the three midpoint tiles and all other tiles blank.
    Bottom: three frames, one per vertex, each holding that vertex's polyomino;
    the polyominoes of u and v are teal (selected) and the polyomino of w is gray (pruned).}
    \caption{The pruning reduction on the triangle $K_3$ ($n = 3$, minimum vertex cover size two).
    \panellabel{Top Left} The input graph with each edge's midpoint marked.
    \panellabel{Top Middle} Its layout on the shared tile grid.
    \panellabel{Top Right} The gap matrix $\bar{G}^{\bar{M}}$:
    every midpoint tile receives gap of $n = 3$,
    forcing the selection of at least one of its edge's two endpoint polyominoes,
    while all other tiles carry no sampling gap ($\bar{g}^{\bar{M}}_{i,j} = \textnormal{null}$, blank).
    \panellabel{Bottom} The three frames,
    one vertex polyomino per frame,
    each padded to $T = 7$ tiles.
    Each midpoint is covered by exactly its edge's two endpoint polyominoes,
    e.g., $m_{(v, w)}$ by $S_v$ and $S_w$ only,
    so a feasible selection must pick an endpoint polyomino per edge,
    i.e., a vertex cover.
    Selecting $S_u$ and $S_v$ (the cover $\{u, v\}$) satisfies all three midpoints at $2T = 14$ tiles,
    and $S_w$ is pruned;
    no single polyomino covers all three midpoints,
    matching $K_3$'s cover size of two.}
    \label{fig:np-reduction-example}
\end{figure}

\paragraph{Step 2: Vertices as polyominoes.}
We next represent each vertex $v$ as a single connected polyomino $S_v$ on this tile grid.
$S_v$ covers
\begin{enumerate}
    \item vertex $v$'s tile,
    \item the midpoint tile $m_e$ of every edge $e$ incident to $v$,
    \item and the edge-path tiles that connect $v$'s tile to each of these midpoints.
\end{enumerate}
$S_v$ thus consists of $v$'s tile at its center and one arm per incident edge;
we call $S_v$ the \emph{vertex polyomino} of $v$.
Since each edge has exactly two endpoints,
each midpoint $m_e$ appears in exactly two vertex polyominoes,
those of $e$'s two endpoints.
Together, the $n$ vertex polyominoes represent the graph:
each vertex $v$ becomes a polyomino $S_v$,
and each edge $(u, v)$ corresponds to its midpoint $m_{(u, v)}$,
the only tile that $S_u$ and $S_v$ share.
The bottom row of \autoref{fig:np-reduction-example} shows the vertex polyominoes of the triangle graph $K_3$.

\paragraph{Step 3: Placing the polyominoes into frames.}
We next spread this representation over time.
We create $n$ video frames,
one frame $f_v$ per vertex $v$,
each its own $R \times C$ tile grid;
the video thus contains $n$ tile grids,
and the single grid of Step~1 is not one of them
but the shared layout from which every frame takes its coordinates.

We place each vertex polyomino $S_v$ alone on its frame $f_v$,
at the same tile coordinates it occupies in the layout of Step~1:
$v$'s tile lies at its layout location,
as does the midpoint $m_{(u, v)}$ of every edge $(u, v)$ incident to $v$.
Each midpoint $m_{(u, v)}$ thus appears at the same tile location in both $f_u$ and $f_v$,
whose polyominoes $S_u$ and $S_v$ cover it.

In the original problem's data model,
polyominoes on the same frame must be disjoint (\autoref{sec:dm-polyominoes}),
so $S_u$ and $S_v$,
which share the midpoint tile $m_{(u, v)}$,
could not coexist on a single frame;
spreading the polyominoes over time is what lets two of them cover the same tile location.

\paragraph{Step 4: Padding polyominoes to a common size.}
The vertex polyominoes generally differ in size,
since arm lengths depend on the layout and on vertex degrees,
so we pad every vertex polyomino to a common size $T$,
the size of the largest polyomino,
by appending \emph{filler} tiles to it.
Filler tiles may lie on any tile except a midpoint,
as long as they keep $S_v$ connected;
avoiding midpoints ensures that each midpoint stays covered by exactly two polyominoes.
All $n$ vertex polyominoes then have the same size $T$,
e.g., $T = 7$ in \autoref{fig:np-reduction-example}.

\paragraph{Step 5: Setting the sampling gaps.}
Finally, we set the maximum tile sampling gaps.
We set $\bar{g}^{\bar{M}}_{i,j} = n$ for all midpoint tiles $(i, j)$,
requiring each to be covered by a selected polyomino at least once across the $n$ frames,
and set $\bar{g}^{\bar{M}}_{i,j} = \textnormal{null}$ on all other tiles,
imposing no requirement anywhere else.
In \autoref{fig:np-reduction-example},
the three midpoint tiles (orange) receive gap of $n = 3$.

\paragraph{Step 6: Equivalence to Minimum Vertex Cover.}
Because the instance contains exactly one polyomino per vertex,
selecting polyominoes is the same as selecting vertices in $G$.
Since only the polyominoes $S_u$ and $S_v$ of edge $(u, v)$'s two endpoints cover $m_{(u, v)}$,
satisfying $m_{(u, v)}$'s sampling-gap requirement means selecting at least one of $S_u$ or $S_v$,
so covering all midpoints is equivalent to selecting a vertex cover.

Because all vertex polyominoes have the same size $T$,
minimizing the total number of selected tiles is equivalent to minimizing the number of selected polyominoes,
which is equivalent to minimizing the number of selected vertices.
The construction is polynomial,
so a polynomial-time algorithm for this pruning variant would solve Minimum Vertex Cover,
and the variant is therefore NP-hard.
\autoref{fig:np-reduction-example} illustrates the reduction on a triangle graph.

\subsubsection{Polyomino Pruning Encodes the Relaxed Variant}
\label{sec:np-encodes-relaxed}
We now adapt the reduction to polyomino pruning as formulated in \autoref{sec:prune},
where every tile carries a sampling gap.

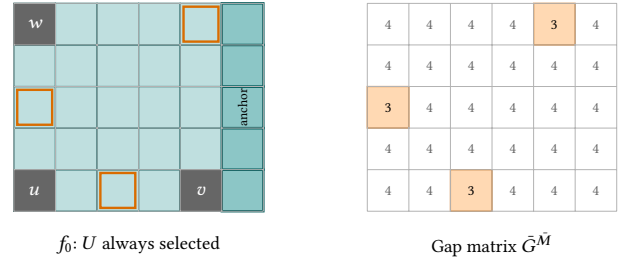
\begin{figure}[t]
    \centering
    \begin{tikzpicture}[x=0.55cm,y=0.55cm]
  \tikzstyle{labelstyle}=[font=\footnotesize]
  \tikzstyle{tinylabel}=[font=\tiny]
  \tikzstyle{guidestyle}=[line width=0.3pt, draw=black!30]
  \tikzstyle{verttile}=[draw=black!70, fill=black!60, line width=0.3pt]
  \tikzstyle{keeptile}=[draw=teal!70!black, fill=teal!25, line width=0.45pt]
  \tikzstyle{anchortile}=[draw=teal!70!black, fill=teal!45, line width=0.9pt]
  \tikzstyle{midframe}=[draw=orange!85!black, fill=none, line width=0.9pt]
  \tikzstyle{midtile}=[draw=orange!85!black, fill=orange!30, line width=0.45pt]

  \foreach \x in {0,...,5} {
    \foreach \y in {0,...,4} {
      \draw[keeptile] (\x,\y) rectangle ++(1,1);
    }
  }
  \foreach \y in {0,...,4} {
    \draw[anchortile] (5,\y) rectangle ++(1,1);
  }
  \node[tinylabel, rotate=90] at (5.5,2.5) {anchor};
  \foreach \x/\y in {2/0,0/2,4/4} {
    \draw[midframe] (\x+0.08,\y+0.08) rectangle ++(0.84,0.84);
  }
  \foreach \x/\y/\n in {0/0/u,4/0/v,0/4/w} {
    \draw[verttile] (\x,\y) rectangle ++(1,1);
    \node[labelstyle, text=white] at (\x+0.5,\y+0.5) {$\n$};
  }
  \draw[guidestyle, step=1] (0,0) grid (6,5);
  \node[labelstyle] at (3,-0.8) {$f_0$: $U$ always selected};

  \begin{scope}[shift={(8.5,0)}]
    \foreach \x/\y in {2/0,0/2,4/4} {
      \draw[midtile] (\x,\y) rectangle ++(1,1);
      \node[tinylabel] at (\x+0.5,\y+0.5) {$3$};
    }
    \foreach \x/\y in {0/0,1/0,3/0,4/0,5/0,
                       0/1,1/1,2/1,3/1,4/1,5/1,
                       1/2,2/2,3/2,4/2,5/2,
                       0/3,1/3,2/3,3/3,4/3,5/3,
                       0/4,1/4,2/4,3/4,5/4} {
      \node[tinylabel, text=black!60] at (\x+0.5,\y+0.5) {$4$};
    }
    \draw[guidestyle, step=1] (0,0) grid (6,5);
    \node[labelstyle] at (3,-0.8) {Gap matrix $\bar{G}^{\bar{M}}$};
  \end{scope}
\end{tikzpicture}
    \Description{Left: a tile grid of five rows and six columns showing the universal polyomino U for the triangle example.
    U covers every tile of the grid,
    including a darker anchor column on the right that no vertex polyomino covers.
    Vertex tiles and midpoint outlines are marked for orientation.
    Right: the gap matrix as the same grid with the value 3 on the three midpoint tiles and 4 on every other tile.}
    \caption{The universal polyomino $U$ from \autoref{sec:np-encodes-relaxed},
    shown for the triangle example of \autoref{fig:np-reduction-example}.
    \panellabel{Left} $U$ occupies its own frame $f_0$ ($N = n + 1 = 4$),
    prepended to the vertex frames,
    and covers the entire grid.
    Every frame's grid is extended by the anchor column (darker);
    no vertex polyomino covers its tiles,
    so they are covered by $U$ alone.
    Vertex and midpoint tiles are marked for orientation only.
    \panellabel{Right} The gap matrix $\bar{G}^{\bar{M}}$:
    midpoint tiles (orange) keep their gap of $n = 3$,
    and every other tile,
    including the anchor column,
    receives gap of $N = 4$.
    The anchor tiles' windows can be satisfied only by $U$,
    forcing $U$ into every feasible selection;
    $U$ then satisfies the windows of all non-midpoint tiles,
    while each midpoint (outlined) is still forced by its vertex-frames window to select an endpoint polyomino.}
    \label{fig:np-universal}
\end{figure}

\paragraph{Step 1: Adding a universal polyomino.}
We call the original $n$ frames,
one per vertex,
the \emph{vertex frames},
and add one new frame $f_0$ before the first of them,
increasing the number of frames from $n$ to $N = n + 1$ (\autoref{fig:np-universal}).
We also extend the tile grid of every frame by one extra column on the right,
which we call the \emph{anchor} column;
the vertex polyominoes lie entirely within the original grid,
so no vertex polyomino covers any anchor tile.
Frame $f_0$ holds a single \emph{universal} polyomino $U$ covering the entire extended grid.
The anchor column's tiles are thus covered by $U$ alone.

\paragraph{Step 2: Setting the sampling gaps on every tile.}
We give every non-midpoint tile,
including the anchor column's tiles,
gap $N$,
and every midpoint tile gap $n$.
Every tile now carries a sampling gap,
as the original pruning problem requires,
and no gap exceeds $N$,
the number of frames.

\paragraph{Step 3: Equivalence to the relaxed variant.}
A non-midpoint tile carries gap $N$,
and a video of $N$ frames contains exactly one window of $N$ consecutive frames,
namely the full video.
The anchor column's tiles are covered only by $U$,
so their windows can be satisfied only by selecting $U$,
and every feasible selection must include $U$.
Selecting $U$ in turn satisfies every non-midpoint tile,
since $U$ covers every tile.
A midpoint tile $m_{(u, v)}$ of an edge $(u, v)$ carries gap $n$,
and the $N = n + 1$ frames contain exactly two windows of $n$ consecutive frames.
One window spans only the $n$ vertex frames;
we call it the \emph{vertex-frames window}.
It still contains just $S_u$ and $S_v$ covering $m_{(u, v)}$,
so it still forces an endpoint polyomino per edge.
The other window contains $U$ and is already satisfied.

A feasible selection thus always consists of $U$ plus a set of vertex polyominoes,
costing $|U|$ plus $T$ per selected vertex,
so the correspondence to vertex covers holds as before,
and polyomino pruning is NP-hard even when every tile carries a gap of at most $N$.

\subsection{Polyomino Packing is NP-hard}
\label{sec:np-packing}
Two-dimensional rectangular bin packing is a well-known NP-hard problem~\cite{garey1979computers,johnson1974np-polyomino,jansen2013np-int-polyomino,lodi2002binpacking2d},
and our polyomino packing problem strictly generalizes it.
Since rectangles are a special case of polyominoes,
any instance of 2D rectangular bin packing can be expressed as an instance of polyomino packing.
Therefore, polyomino packing is at least as hard as 2D rectangular bin packing.

\end{document}